\pdfoutput=1
\documentclass[11pt]{article}
\usepackage[preprint]{latex/acl}
\usepackage{amsmath}
\usepackage{times}
\usepackage{latexsym}
\usepackage{graphicx}
\usepackage[T1]{fontenc}
\usepackage{algorithm}
\usepackage{algorithmic}
\usepackage[utf8]{inputenc}
\usepackage{multirow}
\usepackage{microtype}
\usepackage{hyperref}
\usepackage{inconsolata}

\title{SARA: Singular-Value Based Adaptive Low-Rank Adaption}

\author{Jihao Gu \and Shuai Chen \and Zelin Wang \and Yibo Zhang \and Ping Gong \\
         Beijing University of Posts and Telecommunications \\
  gujihao@bupt.edu.cn}

\begin{document}
\maketitle
\begin{abstract}
% With the increasing cost of fine-tuning large pre-trained models, the importance of parameter-efficient fine-tuning(PEFT) methods becomes more and more prominent. Among these methods, we focus on LoRA, which involves parallel trainable parameters in the multi-head attention part, and has shown promising results. In previous works, research on the mechanistic interpretability of transformer structure may have been overlooked, since the PEFT method is based on the this architecture. According to this insight, we propose the InterLoRA, which integrates LoRA with feature adaptation mechanism into both the attention layer, considering the varying importance of multiple heads, and the FFN layer, acknowledging the memory storage characteristics. Experiments on various complex generation tasks demonstrate the effectiveness of InterLoRA in jointly fine-tuning both parts and controlling parameter memory.

With the increasing number of parameters in large pre-trained models, 
% parameter-efficient fine-tuning methods become more and more important. Among these, the LoRA method is widely used for not adding inference overhead. 
LoRA as a parameter-efficient fine-tuning(PEFT) method is widely used for not adding inference overhead. 
The LoRA method assumes that weight changes during fine-tuning can be approximated by low-rank matrices. However, the rank values need to be manually verified to match different downstream tasks, and they cannot accommodate the varying importance of different layers in the model. 
In this work, we first analyze the relationship between the performance of different layers and their ranks using SVD. Based on this, we design the \textbf{S}ingular-Value Based \textbf{A}daptive Low-\textbf{R}ank \textbf{A}daption(SARA), which adaptively finds the rank during initialization by performing SVD on the pre-trained weights. 
Additionally, we explore the Mixture-of-SARA(Mo-SARA), which significantly reduces the number of parameters by fine-tuning only multiple parallel sets of singular values controlled by a router. 
Extensive experiments on various complex tasks demonstrate the simplicity and parameter efficiency of our methods.
They can effectively and adaptively find the most suitable rank for each layer of each model. 
% These methods can achieve even better performance with fewer parameters.
\end{abstract}

\section{Introduction}
\begin{figure}[!h]
\begin{center}
\includegraphics[scale=0.098]{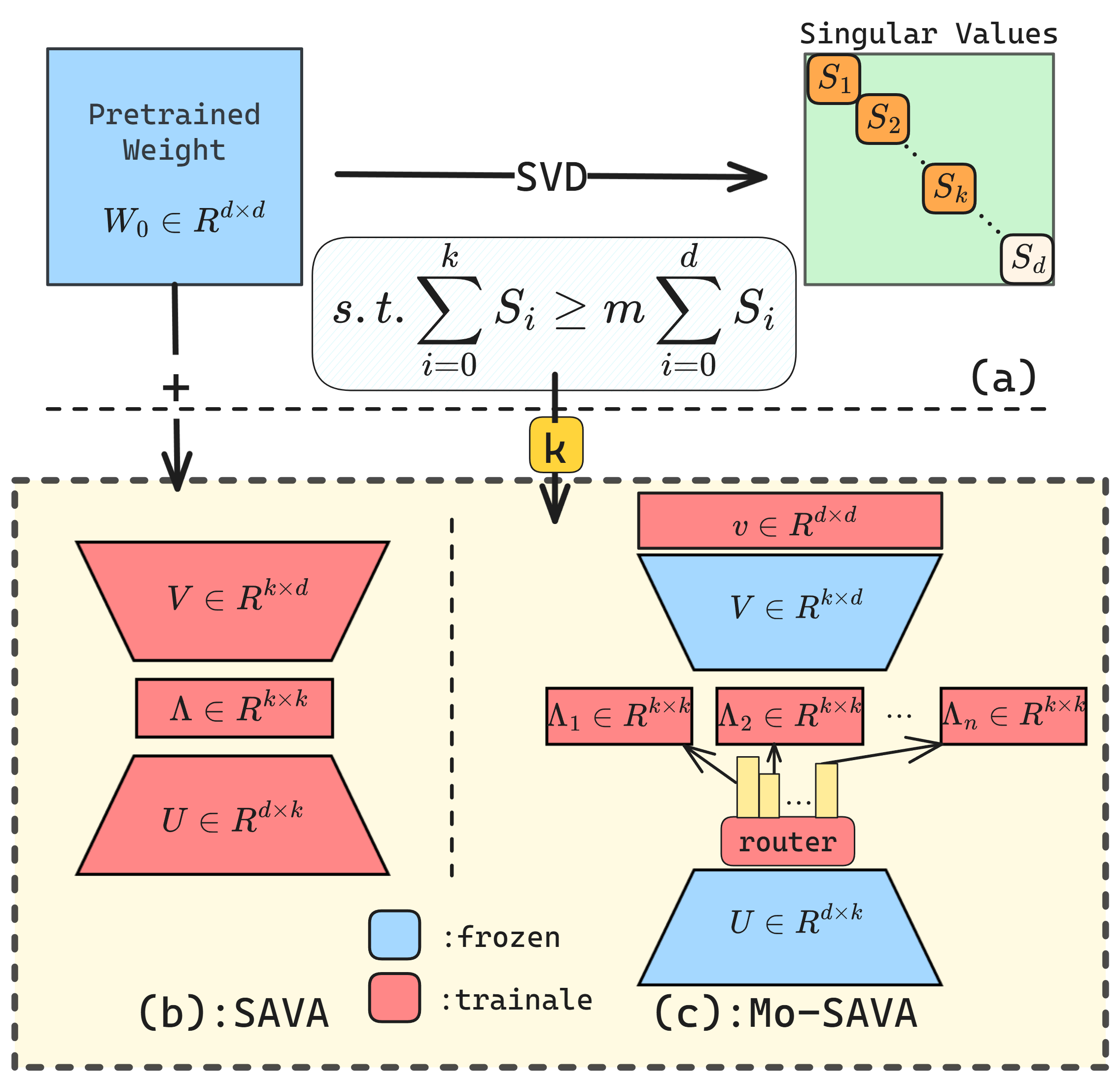}
\caption{An overview of our methods, (a) performing SVD on the pre-trained weights and determining the number $k$ of values that account for a proportion threshold $m$ of the total sum of singular values; (b) the method of adding a truncated singular value matrix to the pre-trained weights based on $k$; and (c) the extreme method of fine-tuning only mixture of parallel singular values. $\Lambda$ and $v$, as diagonal matrix, only require a one-dimensional vector for storage.}
\label{fig.1}
\end{center}
\end{figure}
Large pre-trained language models have demonstrated impressive generative capabilities, achieving excellent performance across various natural language processing(NLP) tasks \citep{touvron2023llama,qin2023chatgpt,kojima2022large}.
However, as the model size increases, the cost of full-parameter fine-tuning to adapt the model to downstream tasks becomes increasingly prohibitive.
To address this issue, Parameter-Efficient Fine-Tuning (PEFT) methods have garnered increasing attention\cite{houlsby2019parameter,li-liang-2021-Prefix}. Among them, the LoRA\cite{hu2021lora} method, which leverages the concept of matrix 'intrinsic rank' by freezing the original model parameters and fine-tuning only a small number of newly added, representative parameters, has been widely adopted. Its primary advantage is that it does not add extra computational overhead during inference.

However, this approach requires setting a uniform rank $r$ as a fixed hyperparameter globally, which not only fails to determine the most suitable rank for each specific model (as the performance is very sensitive to this) but also mismatches with the characteristic that different layers in transformers have varying degrees of importance\citep{jawahar-etal-2019-bert,47786,jawahar-etal-2019-bert}. 
% Meanwhile, the setting of the scaling factor for the LORA method can also lead to different performance
% Under this method, ranks differing by several times can still achieve similar fine-tuning results.
These issues necessitate research into identifying more 'important' parts of the model to find the rank $r$ for fine-tuning.
%and further reducing the number of trainable parameters.

Given that the concept of rank is related to singular value decomposition(SVD), where the decomposed singular diagonal value matrix is characterized by a small number of leading values accounting for a large proportion of the total sum, we perform SVD on the pre-trained model matrices, as shown in Figure \ref{fig.1}a. We conduct experimental analysis on the number $k$ of values in the singular value that cumulatively account for a certain threshold of the total sum. Our analysis reveals that the $k$ value of each layer in the model correlates with the performance of that layer!

Based on this finding, we believe that $k$ reflects the most important part of the parameters for fine-tuning the matrix at the given importance proportion. Then, we propose a \textbf{S}ingular-Value Based \textbf{A}daptive Low-\textbf{R}ank \textbf{A}daption(SARA) method, as shown in Figure \ref{fig.1}b. SARA calculates the most suitable rank for each layer based on the importance threshold during initialization and fine-tunes the newly added truncated singular value matrices. Additionally, we explore an extreme method, Mixture-of-SARA(Mo-SARA), which significantly reduces the number of trainable parameters to the limit. As shown in Figure \ref{fig.1}c, Mo-SARA only fine-tunes $k$ diagonal values as well as a diagonal matrix$v$ used to accelerate convergence. They just require a one-dimensional vector for storage to significantly reduce the number of trainable parameters. Moreover, leveraging the concept of Mixture-of-Experts(MoE) \cite{jacobs1991adaptive}, we innovatively train multiple singular value matrices in parallel, to leverage the entire truncated singular value matrix separately, achieving comparable performance.

% We also validate, from a mathematical analysis perspective, adding singular values for fine-tuning allows for a more fine-grained determination of the LoRA's appropriate scaling factor.

Experimental results show that our improved method can adaptively find more efficient ranks, achieving better performance even with fewer trainable parameters while retaining the advantages of LoRA. 
% and from a mathematical analysis perspective, adding singular values allows for a more fine-grained determination of the appropriate scaling factor.

In summary, we make the following three contributions:

1.We analyze the interactions between different layers and the connections in pre-trained parameter matrices by SVD, discovering more important parameter components. This provides a new research perspective for the entire PEFT field to address the issue of inter-layer information inconsistency.

2.We propose the SARA method, which can adaptively calculate the appropriate rank for each layer during initialization. This method extends the performance of LoRA, inherits all its advantages, and can be combined with other improvement methods.

3.We further propose the Mo-SARA method, which explores leveraging the entire SARA process with only singular values within SARA and paralleling these values. This significantly enhances the efficiency of parameter utilization in PEFT methods.

\section{Related Works}
\subsection{PEFT Methods}
Traditional PEFT methods primarily focus on freezing the original pre-trained model parameters and fine-tuning only a subset of newly added parameters. There are three classic approaches.
These include methods represented by adapters\citep{houlsby2019parameter,patel-etal-2021-nlp}, which involve serially connecting a set of newly added tunable parameters within the model; methods represented by prefix-tuning, which add virtual tokens to the model inputs\citep{li-liang-2021-Prefix}; and methods represented by LoRA\citep{hu2021lora}, which assume that the model parameter matrix only requires fine-tuning a matrix of rank $r$, and replace the original matrix with two matrices that increase and decrease dimensions, respectively, for fine-tuning.

Among these methods, LoRA is widely used while it can be directly added alongside the original matrix without requiring additional inference time and generally achieves better performance across various tasks.
Consequently, the LoRA method has numerous improvements.

\subsection{LoRA's Variants}
AdaLoRA\cite{zhang2023adaptive} and DyLoRA\cite{valipour-etal-2023-dylora} calculate suitable rank during training; Dora\cite{liu2024dora} improves performance by decomposing the original matrix into weight and direction components and fine-tuning them separately; the VeRA\cite{kopiczko2023vera} method reduces the number of trainable parameters based on the LoRA by randomly initializing and freezing the up-sampling and down-sampling matrices and only fine-tuning two diagonal matrices added after them.

Compared to the aforementioned improvements, our main advantage is that we find a simple but efficient enough method to adaptive reveal more important parameter components and find the most suitble rank layer-by-layer through SVD. We also propose a MoE-like method, which leverages a larger number of parameters for fine-tuning with a minimal parameter count. These methods only require some computation time during initialization and retain all the advantages of LoRA, and for the second method, only a minimal number of parameters need to be stored.

\section{Correlation Analysis Between Layer Performance and Singular Values}\label{section3}
As mentioned above, different layers store different importance of information and for the LoRA method, each model adapts to downstream tasks with different ranks.

\begin{figure}[!h]
\begin{center}
\includegraphics[scale=0.28]{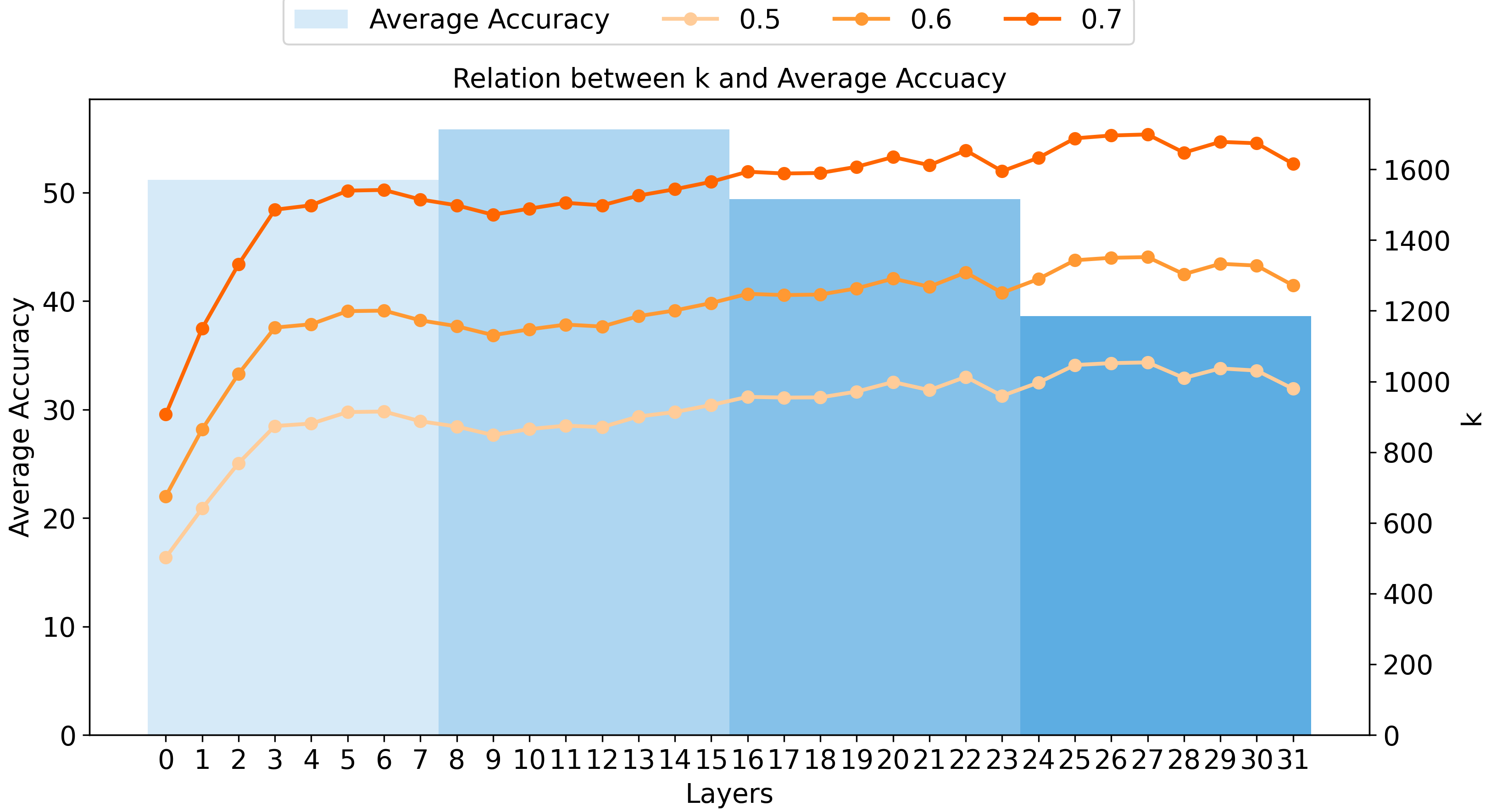}
\caption{The impact of different layers on the average accuracy of mathematical reasoning tasks and the $k$ (mean value obtained from Q and V matrix SVD.)}
\label{fig.2}
\end{center}
\end{figure}

To study the inter-layer importance of LoRA, we conduct a case study experiment on LLaMA-7B\citep{touvron2023llama} based on previous work\cite{hu2023llm}. We divide 32 layers of the model into four parts and fine-tune each part separately using LoRA method, testing their average accuracy on six mathematical reasoning datasets. 
% Specific hyperparameters can be found in the appendixA. 
As shown in the bar chart in Figure\ref{fig.2}, the overall performance is excellent in the lower layers and poorer in the upper layers.

Since the rank concept is related to SVD, we perform SVD on the Q and V matrices used in the classical LoRA method and analyze the singular values. Because the decomposed singular values are arranged in descending order and a small proportion of the leading values account for a large portion of the total sum, we calculate the number of singular values $k$ needed to account for various proportion thresholds. (specific details on obtaining $k$ can be referenced to algorithm\hyperref[alg:calculate-k-value]{1}.) As shown in the line chart in Figure \ref{fig.2}, under all different proportion choices, the value of $k$ decreases initially as the model goes from lower layers to higher layers and then increases, which is exactly opposite to the trend of performance change across layers.

We believe that this is because, to achieve similar effects, the lower layers require a lower 'intrinsic rank' while the upper layers require a higher one. This corresponds to our calculated $k$ values. 
Therefore, allocating the same rank to all layers leads to shortcomings in certain layers, thus affecting the overall efficiency of the model, and it is necessary to allocate ranks to each layer according to the corresponding $k$ values to avoid the bottleneck effect. Based on this, we design an improved method and conduct tests to compare the effects between layers, which will be presented in Section \ref{section5.4}.

% We also discover different behaviors in the Q and V matrices and conduct experiments to demonstrate that our adaptive method can provide the most suitable rank not only for each layer but also for each weight matrix. This will be presented in Appendix A.1.

\section{Method}
\subsection{Motivation}
Based on the above findings, we define the number of singular values accounting for a certain proportion of the total sum after matrix decomposition as $k$, and believe it can be used to reflect the intrinsic rank $r$. Specifically, we use importance proportion threshold instead of rank to set the hyperparameters, adding a new truncated singular value matrix parallel to the original matrix.
% This method only requires some computational time during initialization but can identify the most suitable rank, fully unleashing the potential of the LoRA method.

\subsection{SARA}
The LoRA\cite{hu2021lora} method is based on the assumption that changes in the matrix during fine-tuning have a low 'intrinsic rank' and it involves adding a up-sampling matrix $B\in R^{r\times d}$ and an down-sampling matrix $A\in R^{d\times r}$ with a fixed scaling $\lambda$ parallel to the original weight matrix$W_0\in R^{d\times d}$, using these as the only trainable matrices. The calculation formula is as follows:
\begin{equation}
h =  x(W_0 + \Delta W) = x(W_0 + \lambda \underline{AB})\label{(1)}
\end{equation}

SVD decomposes a matrix into three parts, as shown in Equation \ref{(2)}. 
\begin{equation}
W =  U\Lambda V \approx U_k\Lambda_k V_k \label{(2)}
\end{equation}
The $U\in R^{d\times d}$ and $V\in R^{d\times d}$ are the left and right singular value matrices, respectively. Matrix $\Lambda \in R^{d\times d}$ is called the singular diagonal value matrix with non-negative singular values on the diagonal, arranged in descending order. A small proportion of the leading values accounts for a large portion of the total sum of the singular values. Therefore, a truncated singular value matrix is commonly used to approximate and reduce the original matrix.

In this way, $U_k = U[:,:k]$, $\Lambda_k = \Lambda_k[:k,:k]$, $V_k = V_k[:k,:]$, where $k<d$ needs to be determined in advance.
% $U_k\in R^{d\times k}$,$\Lambda_k \in R^{k\times k}$,$V_k \in R^{k\times d}$

In our methods, we further explore the 'intrinsic rank' from the pre-trained weights of the original matrix using SVD during initialization and use a randomly initialized truncated singular value matrix to represent the part of the original matrix that needs to change during fine-tuning, adding it parallel to the original matrix. The calculation formula is shown as follows:
\begin{equation}
h =  x(W_0 + \underline{U_k\Lambda_k V_k}) \label{(3)}
\end{equation}
where the underlined part represents the trainable truncated singular value matrix, and the calculation of $k$ is shown in the following algorithm\hyperref[alg:calculate-k-value]{1}:

\begin{algorithm}
\caption{Calculate k Value}
\begin{algorithmic}[1]
\REQUIRE $W_{pretrain}\in R^{d\times d}$, $threshold\in (0,1)$
\STATE $U,\Lambda,V \leftarrow \text{SVD}(W_{pretrain})$
\STATE $total \leftarrow \sum \Lambda$
\STATE $target \leftarrow threshold \times total$
\STATE $cumulative \leftarrow 0$, $k \leftarrow 0$
\WHILE{$cumulative < threshold$}
    \STATE $cumulative \leftarrow cumulative + s[k]$
    \STATE $k \leftarrow k + 1$
\ENDWHILE
\STATE \textbf{return}($k$) 
\label{(alg:calculate-k-value)}
\end{algorithmic}
\end{algorithm}

Since the singular value matrix is a diagonal vector, we only need to store a one-dimensional vector, which allows us to square the reduction in parameter storage.

The magnitude of the singular values can indicate the significance of the data. Therefore, we remove the scaling part $\lambda$ in the original LoRA method, as our singular values effectively act as more fine-grained, learnable scaling factors. For $\Delta W$ in SARA, each element is expressed as shown in Equation \ref{(4)}:
\begin{equation}
\Delta W_{ij} \approx \sum_{r=1}^{k} u_{ir} \lambda_{r} v_{rj} \label{(4)}
\end{equation}

\subsection{Mo-SARA}
In singular values, the larger singular values correspond to the main directions of variation in the data, while the smaller singular values can be regarded as noise or less important variations. 
Based on this, we believe that for different downstream tasks, it is sufficient to only adjust the singular values under the same eigenvector mappings. 
Additionally, multiple singular value diagonal matrices can be trained in parallel and selected through a routing mechanism to learn different tasks. 
Therefore, we explore an extreme improvement method for the trainable parameters, called Mixture-of-SARA(Mo-SARA). 
In this method, we keep the left and right singular vectors of the computed truncated singular value matrix unchanged and randomly initialize multiple singular value diagonal matrices for learning. 
To accelerate convergence, we also add a diagonal matrix $v \in R^{d\times d}$ initialized to zero after the truncated singular value matrix. 
The formula for this method is shown in Equation \ref{(5)}:
\begin{equation}
h =  xW_0 + xU_k(\underline{\sum g_i\Lambda_{k_i}}) V_k \underline{v} \label{(5)}
\end{equation}
where the gate $g$=[$g_1$,$g_2$...$g_m$] with values in the range (0,1) is computed as follows:
\begin{equation}
g =  softmax(xU_k \cdot \underline{(W_{g_1}W_{g_2})})\label{(6)}
\end{equation}
Here, we use the value of the input $x\in R^{l\times d}$ with a length of $l$ and dimension $d$ after passing through the left singular matrix $U_k$ as the input, and generate token-level gating $g\in R^{l\times m}$through an MLP layer composed of two gating matrices $W_1 \in R^{k\times 1}$ and $W_2 \in R^{1\times m}$. (We use two one-dimensional linear layers to compute the router to minimize the number of trainable parameters while achieving effective results.)

This method only requires the storage of one-dimensional trainable parameters. Even with multiple parallel sets, it still requires only a few parameters, and each set can leverage the singular values to move the entire truncated singular value matrix, obtaining better efficiency.

\begin{table*}
\centering
\fontsize{9.5pt}{10.5pt}\selectfont
\begin{tabular}{p{2cm}p{1.2cm}p{1.2cm}p{1.2cm}p{1.2cm}p{1.2cm}p{1.2cm}p{1.2cm}p{1.2cm}}  
\hline
\textbf{Method}&\textbf{Params(\%)}&\textbf{SVAMP}&\textbf{AQuA}&\textbf{AddSub}&\textbf{MultiArith}&\textbf{SingleEQ}&\textbf{GSM8K}&\textbf{Avg.} \\
\hline
\multicolumn{9}{c}{\textbf{LLaMA-7B}} \\
Prefix &1.2E-1 & 42.50 & 23.53 & 58.23 & 60.00 & 66.67 & 15.91 & 44.47 \\
Adapter &2.9 & 53.50&	23.53&	74.68&	86.36&	75.49&	20.08&	55.61
 \\
Parallel &2.9 & \textbf{63.00}&	\underline{25.49}&	\underline{75.95}&	86.36&	83.33&	23.11&	59.54 \\
LoRA &7.8E-2 & 58.50&	23.53&	\underline{75.95}&	\textbf{92.73}&	\textbf{88.24}& 24.24&	\underline{60.53} \\
\textbf{Mo-SARA} & \textbf{8.5E-3} & 55.00&	23.53&	70.89&	\underline{90.91}&	\underline{87.25}&	\underline{26.14}&	58.95\\
\textbf{SARA} &\underline{7.1E-2} & \underline{60.00}&	\textbf{35.29}&	\textbf{79.75}&	89.09&	84.31&	\textbf{24.62} & \textbf{62.18}
\\
\hline
\multicolumn{9}{c}{\textbf{LLaMA-13B}} \\
Prefix &9.4E-2 & 58.00 & \underline{29.41} & 72.15 & 78.18 & 82.35 & 22.73 & 57.14 \\
Adapter &2.4 & 55.00&	\textbf{31.37}&	73.42&	78.18&	69.61&	17.05&	54.10 \\
Parallel &2.4 & \underline{69.00} &	17.65&	\underline{81.01}&	\underline{93.64}&	86.27&	27.27&	62.47 \\
LoRA &\underline{6.3E-2} & 66.00&	21.57&	\textbf{82.28}&	\textbf{95.45}&	\textbf{89.22}&	\textbf{36.74}&	65.21 \\
\textbf{Mo-SARA} &\textbf{6.9E-3} & 66.50&	25.49&	\textbf{82.28}&	\textbf{95.45}&	\textbf{89.22}&	34.09&	\underline{65.50} \\
\textbf{SARA} &\underline{6.3E-2} & \textbf{71.50}&	27.45&	\underline{81.01}&	\underline{93.64}&	\underline{88.24}&	3\underline{6.36}&	\textbf{66.37} \\
\hline
\multicolumn{9}{c}{\textbf{GPT-J-6B}} \\
Prefix &1.1E-1 & 41.50 & 9.80 & \textbf{67.09} & 75.45 & 71.57 & 9.85 & 45.88 \\
Adapter &1.9 & 43.00&	13.73&	56.96&	76.36&	64.71& 9.85&	44.10 \\
Parallel &2.8 &42.50&	\underline{19.61}&	56.96& 78.18&	66.67&	\textbf{12.88}&	46.13\\
LoRA &7.6E-2 & \underline{47.00}&	5.88&	\underline{65.82}&	72.73&	76.47&	11.36&	46.54 \\
\textbf{Mo-SARA} & \textbf{8.6E-3} & 45.50&	15.69&	64.56&	\textbf{82.73}&	\textbf{78.43}&	11.74&	\underline{49.77}\\
\textbf{SARA} &\underline{7.0E-2} & \textbf{50.50}& 	\textbf{27.45}& 	\underline{65.82}& \underline{79.09}& \underline{74.51}& 	\underline{12.50}& \textbf{51.65} \\
\hline
\end{tabular}
\caption{The results on six different mathematical reasoning datasets. The answer is the accuracy of calculations obtained using the zero-shot learning method on LLaMA-7B/13B, and GPT-J presented in the table.(\textbf{bold}: the best score; \underline{underline}: the second best)}
\label{table.1}
\end{table*}

\section{Experiment}
In this section, We compare our methods with the latest PEFT methods across a wide range of tasks, including mathematical reasoning, commonsense inference, and E2E tasks, covering a total of 15 datasets. This demonstrates the parameter efficiency of our method. Subsequently, we validate our method's ability to address the issue of inconsistent layer importance mentioned in section \ref{section3}. We then conduct ablation experiments to discuss the importance of each component of our method. Next, we perform experiments to examine the parameter sensitivity of our methods and the impact of the number of parallel heads on Mo-SARA. Finally, we show the routing learned by the Mo-SARA method across various tasks, demonstrating the effectiveness of this mechanism. 
The detailed hyperparameter and experimental settings for all experiments in this section can be found in Appendix \ref{A}.

\subsection{Mathematical Reasoning}
We compare our SARA and Mo-SARA methods with four PEFT methods, including LoRA\cite{hu2021lora}, Prefix\cite{li-liang-2021-Prefix}, Adapter\cite{houlsby2019parameter}, and Parallel Adapter(Parallel)\cite{patel-etal-2021-nlp}, using three LLMs: LLaMA-7B/13B\cite{touvron2023llama}, and GPT-J\cite{wang2021gpt}, across six mathematical reasoning sub-tasks which are (1) the SVAMP \citep{patel-etal-2021-nlp}, (2) the AQuA \citep{ling-etal-2017-program} dataset, (3) the AddSub \citep{hosseini-etal-2014-learning} dataset, (4) the MultiArith \citep{roy2016solving} dataset, (5) the SingleEQ \citep{koncel-kedziorski-etal-2015-parsing} dataset, and (6) the GSM8K \citep{cobbe2021training} dataset.
We largely follow the open-source work\citep{hu2023llm} in terms of experiments and hyperparameter settings, combining the six tasks to create a unified training dataset and testing accuracy on each task separately. Leveraging the generative capabilities of large models, we conduct experiments using the zero-shot approach. 
% Specific experimental settings can be found in Appendix \ref{A}. 
To ensure a fair comparison, we adjust the threshold for $k$ in our method during initialization to achieve a similar number of trainable parameters. The table below lists the proportion of trainable parameters to the total parameters for each method.

Table \ref{table.1} shows that our SARA method significantly outperforms various baseline methods across a wide range of models, achieving up to an 11$\%$ improvement over the LoRA method. Additionally, our Mo-SARA method achieves remarkable results with an order of magnitude fewer trainable parameters, even surpassing all baselines on the LLaMA-13B and GPT-J models.

\begin{table*}
\centering
\fontsize{9.5pt}{10.5pt}\selectfont
\begin{tabular}{p{1.8cm}p{1.2cm}p{1cm}p{1cm}p{1cm}p{1cm}p{1cm}p{1cm}p{1cm}p{1cm}p{1cm}}
\hline
\textbf{ Method}& \textbf{Params(\%)}&\textbf{ARC-c}&\textbf{ARC-e}&\textbf{Boolq}&\textbf{WinoG}&\textbf{PIQA}&\textbf{SIQA}&\textbf{OBQA}&\textbf{ HellaS}&\textbf{Avg.} \\
\hline
% \multicolumn{11}{c}{\textbf{ChatGPT}} \\
 \textbf{ChatGPT}& -& 79.9 & 89.8 & 73.1 & 66.1 & 85.4 & 68.5 & 74.8 & 78.5 &77.0 \\
\hline
\multicolumn{11}{c}{\textbf{LLaMA-7B}} \\
Prefix &1.1E-1 & 54.0 & 72.9 & 64.3 & 72.1 & 76.8 & 73.9 & 60.6 & 42.1 &64.6 \\
Adapter &9.9E-1 & 57.1&	74.5&	63.0&	75.7&	79.2&	76.3&	72.4 & 67.9 &70.8 \\
Parallel &3.5 & 57.3 & 73.7 & 67.9 & 78.9 & 76.4& 78.8 & 75.2 & 69.8 & 72.2 \\
LoRA &8.3E-1 & 61.3& 77.8 & 68.9 & 78.8 & 80.7 & 77.4 & 74.8 &78.1& 74.7 \\
DoRA & 8.4E-1 & \textbf{66.2} & \textbf{81.9} & \underline{69.7} & \underline{81.0} & \underline{83.4} & \underline{78.6} & \underline{79.2} & \textbf{87.2} & \underline{78.4} \\ 
\textbf{Mo-SARA} &\textbf{8.5E-3} & 54.5 & 74.5 &	62.8 & 71.8 & 76.0 & 73.8 & 65.8 & 50.3 & 66.2 \\
\textbf{SARA} & \underline{8.3E-1} & \underline{65.8}&	\underline{81.6}&	\textbf{70.9}&	\textbf{82.6}&	\textbf{83.6}&	\textbf{78.8}&	\textbf{81.4}&	\underline{82.9}&	\textbf{78.5}
\\
\hline
\multicolumn{11}{c}{\textbf{LLaMA-13B}} \\
Prefix & 3.1E-2 &  62.9 & 79.5 & 65.3 & 68.6 & 75.4 & 72.1 & 68.0 & 55.2 & 68.4 \\
LoRA &6.7E-1 & 68.3 & 82.8 & 72.1 & 83.5 & 80.5 & \textbf{83.7} & \underline{82.4} & 90.5 & 80.5 \\
Adapter &8.0E-1 & 67.3 & 82.5 & 71.8 & 82.4 & 83 & 79.2 & 81.8 & 88.1 & 79.5 \\
Parallel &2.9 & \textbf{71.2} & \textbf{84.2} & \underline{72.5} & 84.1 & \textbf{84.9} & 79.8 & 82.4 & 92.1 & 81.4 \\
LoRA &6.7E-1 & 68.3 & 82.8 & 72.1 & 83.5 & 80.5 & \textbf{83.7} & \underline{82.4} & 90.5 & 80.5 \\
DoRA & 6.8E-1 & 69.6 & \textbf{84.2} & 72.4 & \underline{84.2} & \textbf{84.9} & \underline{81.5} & 82.8 & \textbf{92.4} & \underline{81.5} \\ 
\textbf{Mo-SARA} &6.9E-3 & 61.6& 78.7& 67.9& 76.9& 80.2& 76.3& 72.6& 76.4& 73.8\\
\textbf{SARA} &6.8E-1 & \underline{69.8} &	\underline{84.1}&	\textbf{73.2}&	\textbf{84.9}&	\underline{83.9}&	80.6&	\textbf{84.6}&	\underline{92.2} &	\textbf{81.7}\\
\hline
\end{tabular}
\caption{\label{citation-guide} The results on 8 commonsense inference datasets, with ChatGPT and baseline results taken from \cite{hu2023llm}, and the DoRA method results sourced from \cite{liu2024dora}.(\textbf{bold}: the best score; \underline{underline}: the second best)}
\label{table.2}
\end{table*}

\subsection{Commonsense Inference}
For the commonsense reasoning task, which includes eight downstream tasks as follows:(1) the ARC-c and (2) the ARC-e are the Challenge Set and Easy Set of ARC \citep{clark2018think}, (3) the Boolq \citep{clark-etal-2019-boolq}, (4) the WinoGrande \citep{sakaguchi2021winogrande}, (5) the PIQA \citep{bisk2020piqa}, (6) the SIQA \citep{sap2019socialiqa},  (7) the OBQA \citep{mihaylov2018can}, and (8) the HellaSwag \cite{zellers2019hellaswag}. We conduct experiments on the LLaMA-7B/13B\cite{touvron2023llama} models to extend the comparison with DoRA\cite{liu2024dora} method and the results obtained with GPT-3.5-turbo API through zero-shot CoT\cite{wei2022chain}. 
We also largely follow this work\citep{hu2023llm}.
% and specific experimental settings can be found in Appendix \ref{A}.

The results in Table \ref{table.2} show that SARA achieves better results across a variety of models and datasets, with up to a 5\% improvement over the LoRA method. Our Mo-SARA method, despite inherently using fewer training parameters, achieves comparable results on this task with almost two orders of magnitude fewer parameters, even surpassing the performance of the prefix method.

\subsection{E2E Benchmark}
To further validate the performance of our method through broader comparisons, we also conduct experiments on the E2E task\cite{novikova2017e2e}. We follow the experimental setup from \cite{hu2021lora} and use GPT-2 Medium\cite{radford2019language} model. In addition to LoRA, we compare new variants of the LoRA method, including Adalora\cite{zhang2023adaptive}, Dylora\cite{valipour2022dylora}, and Vera\cite{kopiczko2023vera}. For VeRA method, we use all the experimental settings mentioned in the paper\cite{kopiczko2023vera}. 
% All specific experimental parameters can be found in Appendix \ref{A}.

The experimental results are shown in Table \ref{table.3}. It can be seen that our SARA method achieves better results with a smaller number of trainable parameters. Additionally, our Mo-SARA method outperforms the VeRA\cite{kopiczko2023vera}, achieving better results with fewer parameters.

\begin{table*}
\centering
\fontsize{9.5pt}{10.5pt}\selectfont
\begin{tabular}{p{1.8cm}p{1.8cm}p{1.8cm}p{1.8cm}p{1.8cm}p{1.8cm}p{1.8cm}}
\hline
\textbf{Method}& Params&\textbf{BLEU}&\textbf{NIST}&\textbf{METEOR}&\textbf{ROUGE-L}&\textbf{CIDEr} \\
\hline
% \multirow{10}{*}{\textbf{Medium}} & 
$FT^1$ & 354.92M&	68.2&	8.62&	46.2&	 71.0&	2.47
 \\
$Adpt^{L1}$ & 0.37M&	66.3&	8.41&	45.0&	69.8&	 2.40 \\
$Adpt^{L1}$ & 11.09M&	68.9&	8.71&	46.1&	 71.3&	 2.47 \\
$Adpt^{H1}$ & 11.09M&	67.3&	 8.50&	46.0&	70.7&	  2.44 \\
$DyLoRA^2$ & 0.39M&	69.2&	8.75&	46.3&	70.8&	  2.46 \\
$AdaLoRA^3$ & 0.38M&	68.2&	8.58&	44.1&	70.7&	   2.35 \\
$LoRA^1$ & 0.35M&	\textbf{70.4}&	\textbf{8.85}&	 \textbf{46.8}&	71.8 &	  2.53 \\
VeRA & 0.098M&	69.1&	8.71 &	 46.3&	70.8 &	  2.43 \\
Mo-SARA & \textbf{0.094M}&	69.4&	8.77&	46.4&	71.1&	  2.48 \\
SARA & 0.33M&	\textbf{70.4}&	8.84&	46.7&	\textbf{72.3}&	  \textbf{2.55} \\
% \hline
% \multirow{7}*{\textbf{large}} & FT^1 & 774.03M&	68.5&	 8.78&	46.0&	 69.9&	2.45
%  \\
% ~ & Adpt & 0.88M&	69.1&	8.68&	46.3&	 71.4&	  2.49 \\
% ~ & Adpt & 23.00M&	68.9&	 8.70&	 46.1&	  71.3&	  2.45 \\
% ~ & LoRA & 0.77M&	70.1&	 8.80&	 46.7&	  71.9&	  2.52 \\
% ~ & VeRA & 0.17M&	70.3&	 8.85 &	 46.9&	  71.6&	  2.54 \\
% ~ & Mo-SARA & 0.17M&	70.3&	 8.85 &	 46.9&	  71.6&	  2.54 \\
% ~ & SARA & 0.17M&	70.3&	 8.85 &	 46.9&	  71.6&	  2.54 \\
\hline
\end{tabular}
\caption{\label{citation-guide} The results on the E2E dataset, with the results for ($^1,^2,^3$) taken from previous work.$^1$\cite{hu2021lora},$^2$\cite{zhang2023adaptive},$^3$\cite{valipour-etal-2023-dylora}}
\label{table.3}
\end{table*}

\subsection{Improvement of SARA across Layers}\label{section5.4}
\begin{figure}[!h]
\begin{center}
\includegraphics[scale=0.3]{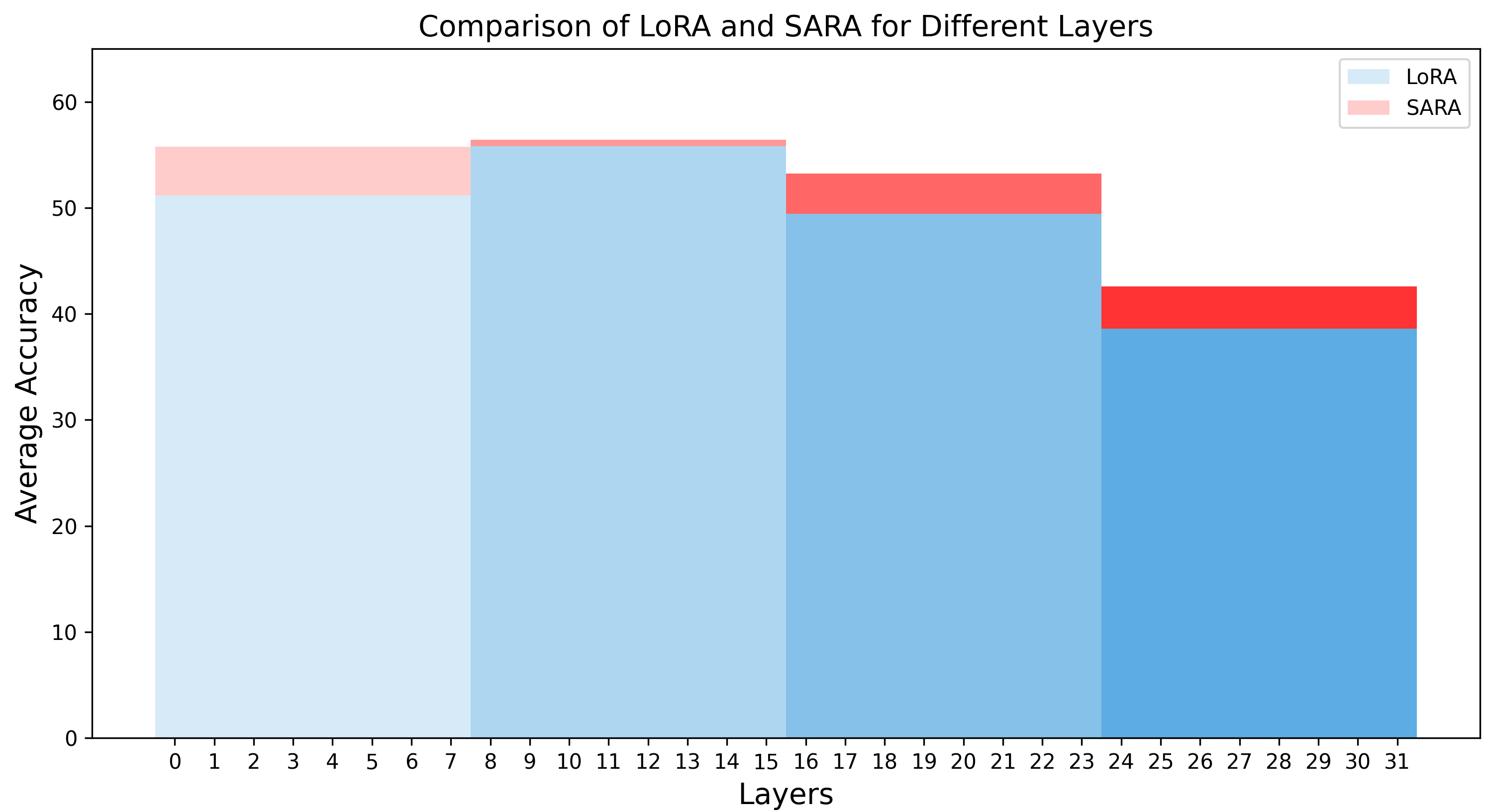}
\caption{Average accuracy of SARA and LoRA methods across layers in mathematical reasoning tasks.}
\label{fig.3}
\end{center}
\end{figure}

We use the SARA method, dividing 32 layers of LLaMA-7B into four parts for separate fine-tuning to verify our method's effectiveness in allocating ranks using singular values, addressing the issue of poorer results caused by inconsistent importance across layers. As shown in the Figure \ref{fig.3}, our method consistently outperforms in each fine-tuning part, reducing the variance among layer results and addressing the problem posed in section \ref{section3}.

% \textbf{Ablation:}On the basis above, we also explore the importance of the internal components of the Mo-SARA method. We try omitting the diagonal matrix $b$, which is added after the singular value matrix for fast convergence, and also placing $b$ before the truncated singular value matrix. The experimental results are shown in Table 4. It can be seen that regardless of its position, a diagonal matrix for fast convergence plays a significant role. At the same time, even when only fine-tuning the singular value part, it still achieves decent results with a small parameter count, proving the importance of this component for fine-tuning, consistent with our hypothesis.

\begin{table*}
\centering
\fontsize{9.5pt}{10.5pt}\selectfont
\begin{tabular}{p{2cm}p{1.2cm}p{1.2cm}p{1.2cm}p{1.2cm}p{1.2cm}p{1.2cm}p{1.2cm}p{1.2cm}}
\hline
\textbf{Method}&Params(\%)&\textbf{SVAMP}&\textbf{AQuA}&\textbf{AddSub}&\textbf{MultiArith}&\textbf{SingleEQ}&\textbf{GSM8K}&\textbf{Avg.} \\
\hline
\textbf{SARA} &7.1E-2 & \textbf{60.00}&	\textbf{35.29}&	\textbf{79.75}&	89.09&	84.31&	24.62 & \textbf{62.18}\\
\hline
\textbf{w/o $\Lambda$} & 7.1E-2 &47.50&	17.65&	65.82&	80.00&	72.55&	12.88&	49.40\\
\textbf{$V$=0} & 7.1E-2 & 58.00 & 17.65 & 74.68 & 90.00 & 86.27 & 26.52 & 58.85 \\
\textbf{w/o $\Lambda$,$V$=0} & 7.1E-2 & \textbf{60.00} & 23.53 & 77.22 & 88.18 & 80.39 & 21.21 &	58.42\\
\hline
\textbf{Mo-SARA} & \textbf{8.5E-3} & 55.00&	23.53&	70.89&	90.91&	\textbf{87.25}&	\textbf{26.14}&	58.95\\
\hline
\textbf{w/o $v$} & 9.0E-3 &49.00&25.49& 69.62& 80.00& 75.49& 18.18& 52.88\\
\textbf{$v$ in front} & \textbf{8.5E-3} & 57.00&	21.57&	73.42&	\textbf{91.82}&	86.27&	23.48&	58.93\\
\hline
\end{tabular}
\caption{\label{citation-guide} Ablation of SARA and Mo-SARA methods on the mathematical reasoning tasks with LLaMA-7B.}
\label{table.4}
\end{table*}
\subsection{Ablation Study}
To analyze the impact of each component of SARA, we set up two groups of ablation experiments. These experiments verify whether it is necessary to initialize the up-projection matrix $V$ to zero as in the original LoRA method and whether it is necessary to add the singular value diagonal matrix $\Lambda$. We conduct experiments using LLaMA-7B on mathematical reasoning tasks, as shown in the table \ref{table.4}.
It can be seen that our approach of directly adding the truncated singular value matrix next to the original matrix yields better results and ddding singular value diagonal matrix almost does not increase the parameter count. 

Additionally, we conduct a set of experiments on the scaling value $\lambda$ of the original LoRA method to show that the original LoRA method is also sensitive to the $\lambda$ and our method of replacing scaling with singular values to some extent addresses this issue. The results can be seen in Appendix \ref{C}.

% \begin{figure}[!h]
% \begin{center}
% \includegraphics[scale=0.45]{scal (2).png}
% \caption{Average accuracy of the LoRA method on mathematical reasoning tasks at different λ scaling ratios compared to the SARA method.}
% \label{fig.1}
% \end{center}
% \end{figure}

% Additionally, we conduct a set of experiments on the scaling value $\lambda$ of the original LoRA method, as shown in Figure 5. The original LoRA method is also sensitive to the $\lambda$, requiring validation for different tasks to determine the most suitable value. Our method of replacing scaling with singular values to some extent addresses this issue.

For Mo-SARA, we try omitting the diagonal matrix $v$ (for a fair comparison of parameter quantities, we parallel 10 heads.), which is added after the singular value matrix for fast convergence, and also placing $v$ in front of the truncated singular value matrix. The experimental results are shown in Table \ref{table.4}. It can be seen that regardless of its position, a diagonal matrix for fast convergence plays a significant role. At the same time, even when only fine-tuning the singular value part, it still achieves decent results with a small parameter count, proving the importance of this component for fine-tuning, consistent with our hypothesis.

\subsection{Robustness of the SARA Method}
\begin{figure}[!h]
\begin{center}
\includegraphics[scale=0.4]{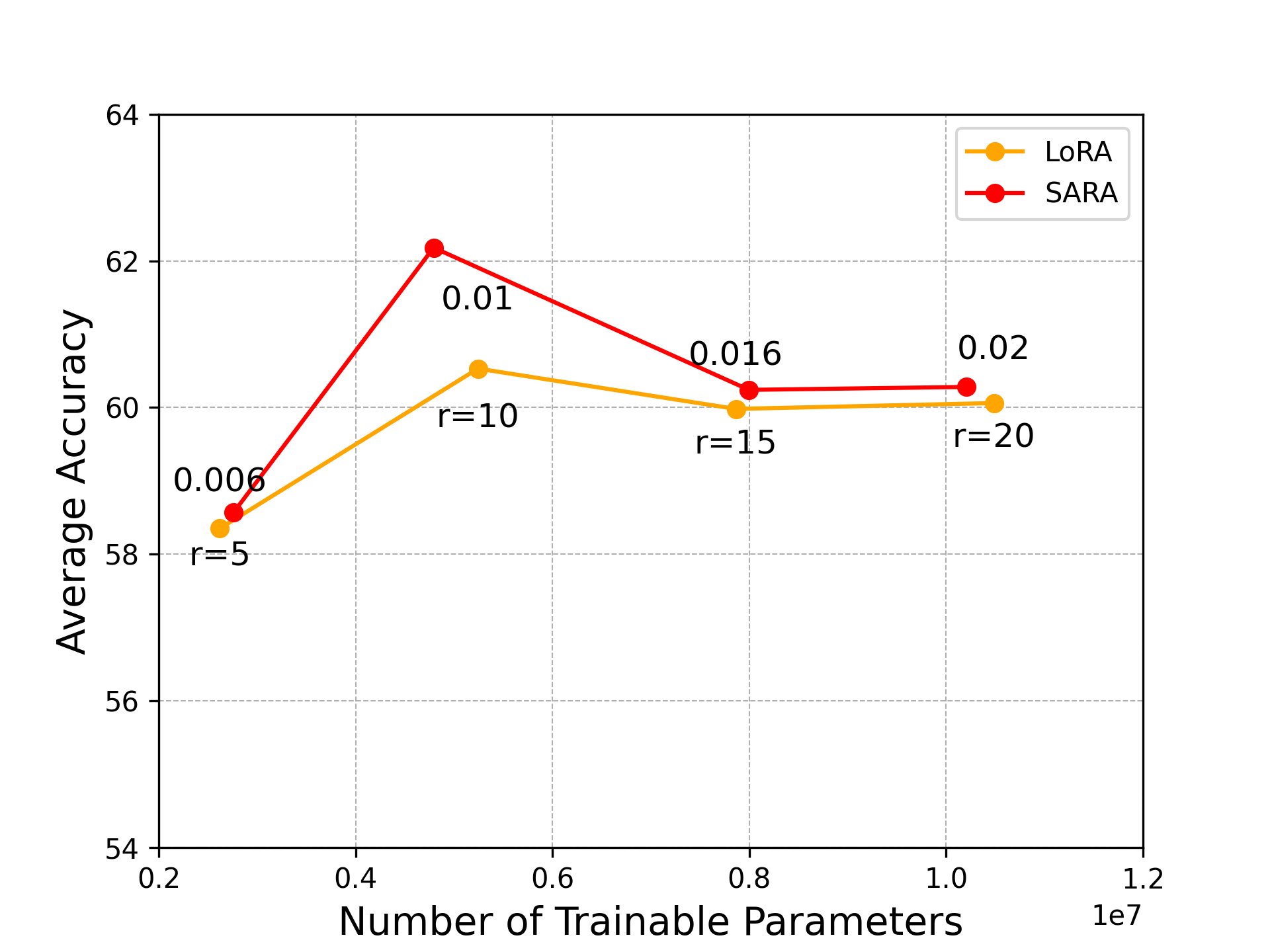}
\caption{Average accuracy of the SARA and LoRA methods on mathematical reasoning tasks with different trainable parameters. The thresholds for determining k in the SARA method [0.006, 0.01, 0.016, 0.02] and the r values used to adjust the parameter count in the LoRA method [5, 10, 15, 20] are indicated in the figure.}
\label{fig.6}
\end{center}
\end{figure}

In this section, we conduct experiments using LLaMA-7B on mathematical reasoning tasks to compare the trends of our method and the LoRA method under different trainable parameter sizes. The experimental results are shown in Figure \ref{fig.6}. It can be seen that our method outperforms the LoRA method under all trainable parameter sizes and exhibits similar trends to the LoRA method.

\subsection{Analysis Under Parameter Limits}
To explore methods for further reducing the number of trainable parameters, we conduct experiments on the Mo-SARA using LLaMA-7B with mathematical reasoning tasks, starting with an experiment on threshold setting for finding the value of $k$, and demonstrating the importance of fine-tuning only the singular values. 
We then analyze the impact of the number of parallel heads, showcasing the role of mixture parallel sets of singular values. 
% Following this, we analyze the effect of adding a diagonal matrix initialized to zero after the truncated singular values in this extreme method.

\begin{figure}[!h]
\begin{center}
\includegraphics[scale=0.4]{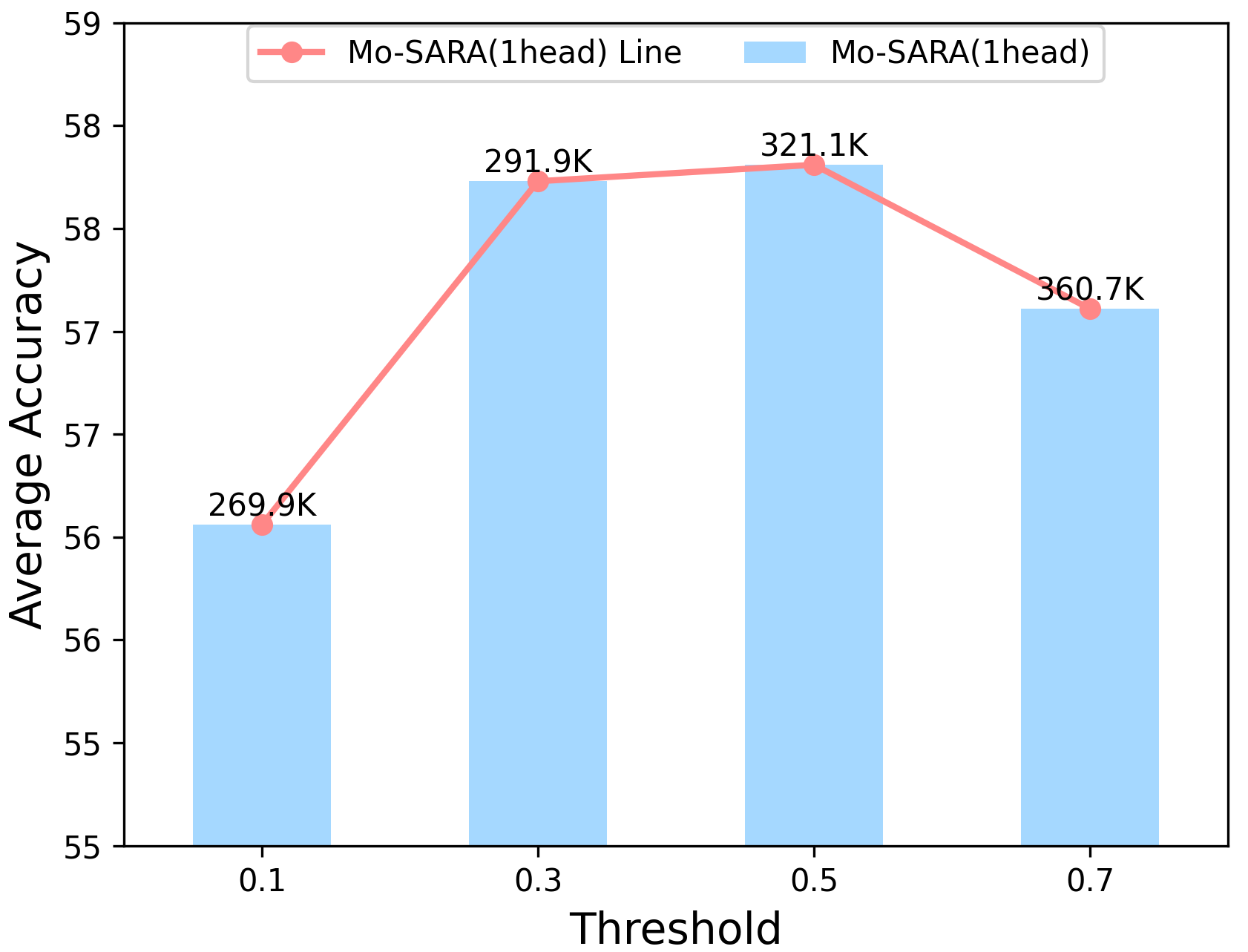}
\caption{Average accuracy of Mo-SARA (1 head) on mathematical reasoning tasks under different thresholds, the bar chart displays the trainable parameters above.}
\label{fig.4}
\end{center}
\end{figure}

\textbf{Threshold:} We design four sets of experiments without parallel structure. The threshold for determining $k$ is set incrementally to 0.1, 0.3, 0.5, and 0.7. The parameter counts and average results across six mathematical reasoning datasets are shown in Figure \ref{fig.4}, displaying a trend of initial increase followed by a gradual decrease. In subsequent experiments, we use 0.5 as the threshold for determining $k$ in the Mo-SARA method.

\begin{figure}[!h]
\begin{center}
\includegraphics[scale=0.4]{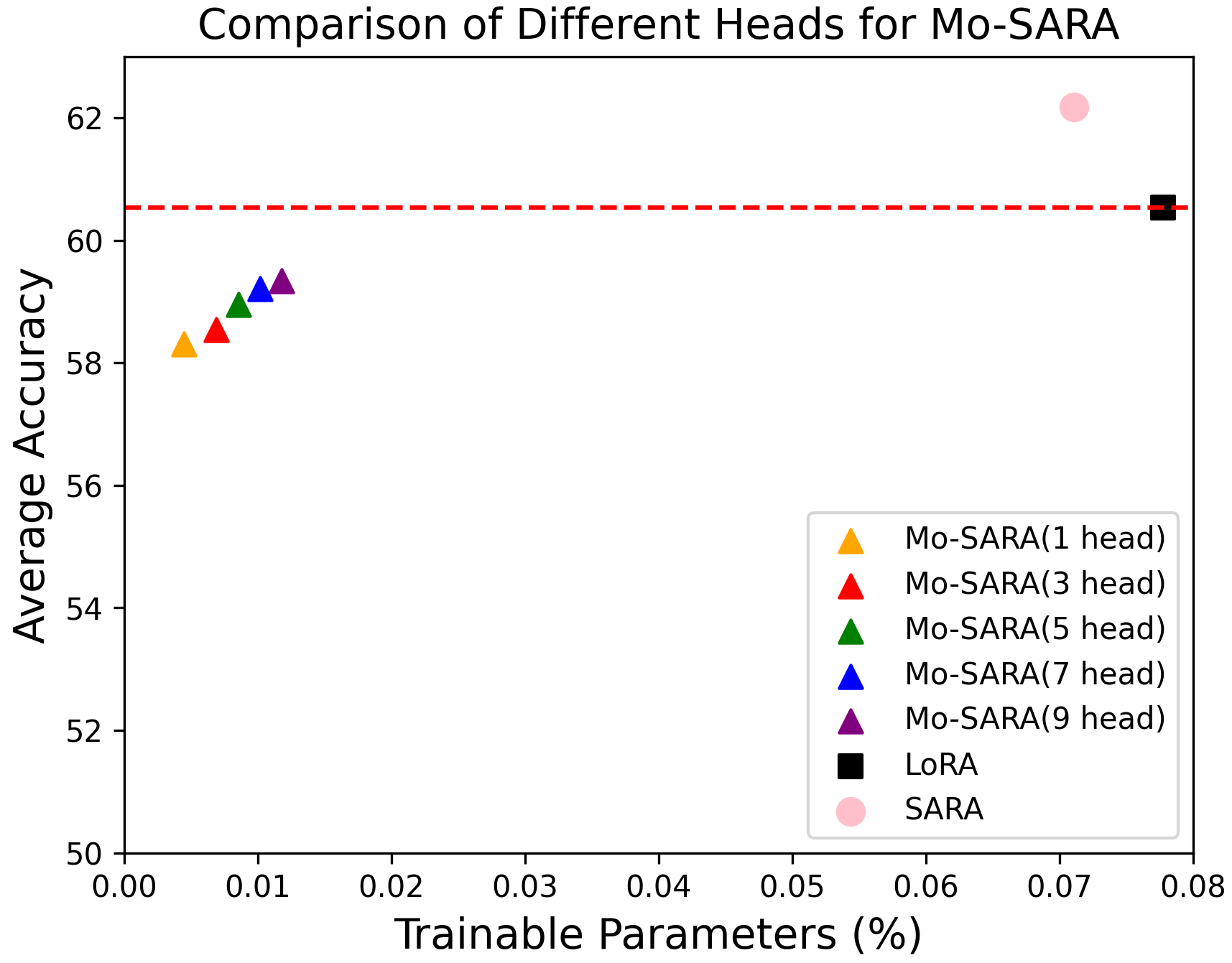}
\caption{Average accuracy of Mo-SARA on mathematical reasoning tasks with different numbers of parallel heads, compared to SARA and LoRA methods.}
\label{fig.5}
\end{center}
\end{figure}

\textbf{Parallel Heads:} We further explore the choice of parallel heads for the parallel structure, using soft routing to control 3, 5, 7, and 9 groups of parallel singular values and compare the results with that without parallel structure. As shown in Figure \ref{fig.5}, the experimental results demonstrate a stable increase in performance as the number of parallel heads increases, gradually approaching the results of the original LoRA method with nearly ten times the parameter count. Considering the balance between parameter count and performance, we adopt a structure with 5 parallel groups in main experiments.

\subsection{Analysis of Routing Effects}\label{section5.8}
\begin{figure}[!h]
\begin{center}
\includegraphics[scale=0.055]{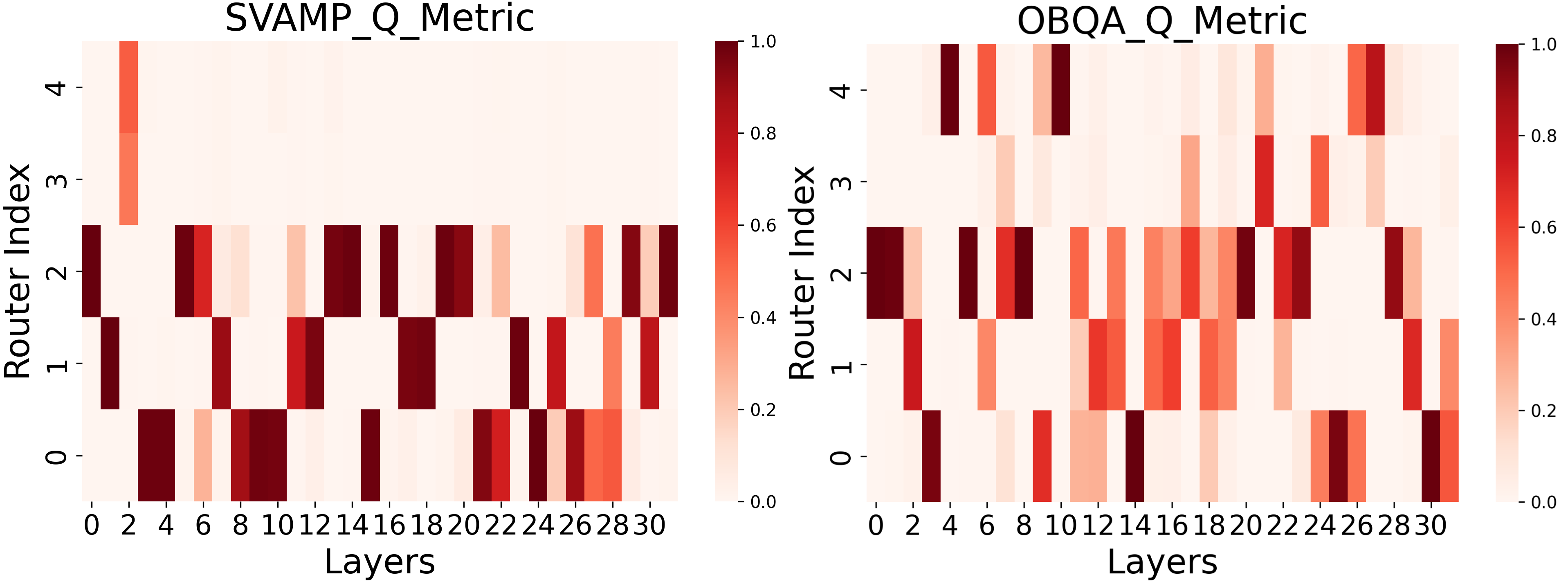}
\caption{The heatmap of routing generated by the model trained with the Mo-SARA on mathematical and commonsense inference tasks through test tasks.}
\label{fig.7}
\end{center}
\end{figure}

To explore the effect of using mixture parallel structure in the Mo-SARA method, we employ the model trained on LLaMA-7B to extract the first question across various test tasks. The routing results of the first model pass are averaged across 'batch' and 'length' dimensions to obtain the routing's heatmap. Figure\ref{fig.7} illustrates the routing results of the Mo-SARA method applied alongside the Q-matrix in mathematical reasoning and commonsense inference tasks. It is observed that for different tasks, the routing mechanism learns different allocation strategies, assigning different weights to each set of singular values, and each of them also learns the tasks it excels at. This indicates the role of the routing in assisting the Mo-SARA method in parallelizing and leveraging the entire singular value matrix for fine-tuning. The remaining test tasks and routing experimental results on matrices are referenced in Appendix \ref{D}.

\section{Conclution}
In this work, we first analyze the relationship between the SVD results of pre-trained model parameters and the performance across layers, identifying the most suitable rank of the matrix and providing a new perspective for addressing the varying importance across layers. 
Based on this, we propose a simple yet efficient method, SARA, which can adaptively find the appropriate rank by calculating the SVD during initialization. We further introduce the Mo-SARA method, which uses a structure that only fine-tunes the routing mechanism and the mixture of singular values it controls, significantly reducing the trainable parameters. Various comparative experiments on 15 datasets demonstrate our methods' higher performance while retaining the advantages of the LoRA method, advancing the field of PEFT by improving performance and further reducing the trainable parameters.

\section{Limitation}
Although our method retains the advantages of the LoRA method, allowing the additional parameter parts to be directly loaded alongside the original matrix without extra inference overhead, there is still a small time cost during training initialization. In the future, we will investigate methods to accelerate SVD decomposition to further speed up our model's training process.
Meanwhile, our proposed Mo-SARA method adopts a mechanism similar to MoE\cite{jacobs1991adaptive}, using a token-level soft routing approach for the gating mechanism, which selects all experts and performs a weighted sum based on the gating. Although we have not conducted extensive research on the choice of gating methods, we have already achieved excellent results as presented. In the future, we will study more MoE methods, to further explore the potential of PEFT methods with minimal parameter sizes.

\section{Ethic Statement}
The main purpose of this paper is to explore effective fine-tuning methods in low-resource scenarios. By using SVD, we investigate the relationship between pre-trained matrices and the performance of different layers in the model, and propose two efficient fine-tuning methods that significantly reduces the number of trainable parameters. All the models and datasets we used are open source, so we believe that the work in this paper does not pose any potential threats.

\bibliography{anthology,custom}

\begin{thebibliography}{32}
\expandafter\ifx\csname natexlab\endcsname\relax\def\natexlab#1{#1}\fi

\bibitem[{Bisk et~al.(2020)Bisk, Zellers, Gao, Choi et~al.}]{bisk2020piqa}
Yonatan Bisk, Rowan Zellers, Jianfeng Gao, Yejin Choi, et~al. 2020.
\newblock Piqa: Reasoning about physical commonsense in natural language.
\newblock In \emph{Proceedings of the AAAI conference on artificial intelligence}, volume~34, pages 7432--7439.

\bibitem[{Clark et~al.(2019)Clark, Lee, Chang, Kwiatkowski, Collins, and Toutanova}]{clark-etal-2019-boolq}
Christopher Clark, Kenton Lee, Ming-Wei Chang, Tom Kwiatkowski, Michael Collins, and Kristina Toutanova. 2019.
\newblock \href {https://doi.org/10.18653/v1/N19-1300} {{B}ool{Q}: Exploring the surprising difficulty of natural yes/no questions}.
\newblock In \emph{Proceedings of the 2019 Conference of the North {A}merican Chapter of the Association for Computational Linguistics: Human Language Technologies, Volume 1 (Long and Short Papers)}, pages 2924--2936, Minneapolis, Minnesota. Association for Computational Linguistics.

\bibitem[{Clark et~al.(2018)Clark, Cowhey, Etzioni, Khot, Sabharwal, Schoenick, and Tafjord}]{clark2018think}
Peter Clark, Isaac Cowhey, Oren Etzioni, Tushar Khot, Ashish Sabharwal, Carissa Schoenick, and Oyvind Tafjord. 2018.
\newblock Think you have solved question answering? try arc, the ai2 reasoning challenge.
\newblock \emph{arXiv preprint arXiv:1803.05457}.

\bibitem[{Cobbe et~al.(2021)Cobbe, Kosaraju, Bavarian, Chen, Jun, Kaiser, Plappert, Tworek, Hilton, Nakano et~al.}]{cobbe2021training}
Karl Cobbe, Vineet Kosaraju, Mohammad Bavarian, Mark Chen, Heewoo Jun, Lukasz Kaiser, Matthias Plappert, Jerry Tworek, Jacob Hilton, Reiichiro Nakano, et~al. 2021.
\newblock Training verifiers to solve math word problems.
\newblock \emph{arXiv preprint arXiv:2110.14168}.

\bibitem[{Hosseini et~al.(2014)Hosseini, Hajishirzi, Etzioni, and Kushman}]{hosseini-etal-2014-learning}
Mohammad~Javad Hosseini, Hannaneh Hajishirzi, Oren Etzioni, and Nate Kushman. 2014.
\newblock \href {https://doi.org/10.3115/v1/D14-1058} {Learning to solve arithmetic word problems with verb categorization}.
\newblock In \emph{Proceedings of the 2014 Conference on Empirical Methods in Natural Language Processing ({EMNLP})}, pages 523--533, Doha, Qatar. Association for Computational Linguistics.

\bibitem[{Houlsby et~al.(2019)Houlsby, Giurgiu, Jastrzebski, Morrone, De~Laroussilhe, Gesmundo, Attariyan, and Gelly}]{houlsby2019parameter}
Neil Houlsby, Andrei Giurgiu, Stanislaw Jastrzebski, Bruna Morrone, Quentin De~Laroussilhe, Andrea Gesmundo, Mona Attariyan, and Sylvain Gelly. 2019.
\newblock Parameter-efficient transfer learning for nlp.
\newblock In \emph{International Conference on Machine Learning}, pages 2790--2799. PMLR.

\bibitem[{Hu et~al.(2021)Hu, Shen, Wallis, Allen-Zhu, Li, Wang, Wang, and Chen}]{hu2021lora}
Edward~J Hu, Yelong Shen, Phillip Wallis, Zeyuan Allen-Zhu, Yuanzhi Li, Shean Wang, Lu~Wang, and Weizhu Chen. 2021.
\newblock Lora: Low-rank adaptation of large language models.
\newblock \emph{arXiv preprint arXiv:2106.09685}.

\bibitem[{Hu et~al.(2023)Hu, Lan, Wang, Xu, Lim, Lee, Bing, and Poria}]{hu2023llm}
Zhiqiang Hu, Yihuai Lan, Lei Wang, Wanyu Xu, Ee-Peng Lim, Roy Ka-Wei Lee, Lidong Bing, and Soujanya Poria. 2023.
\newblock Llm-adapters: An adapter family for parameter-efficient fine-tuning of large language models.
\newblock \emph{arXiv preprint arXiv:2304.01933}.

\bibitem[{Jacobs et~al.(1991)Jacobs, Jordan, Nowlan, and Hinton}]{jacobs1991adaptive}
Robert~A Jacobs, Michael~I Jordan, Steven~J Nowlan, and Geoffrey~E Hinton. 1991.
\newblock Adaptive mixtures of local experts.
\newblock \emph{Neural computation}, 3(1):79--87.

\bibitem[{Jawahar et~al.(2019)Jawahar, Sagot, and Seddah}]{jawahar-etal-2019-bert}
Ganesh Jawahar, Beno{\^\i}t Sagot, and Djam{\'e} Seddah. 2019.
\newblock \href {https://doi.org/10.18653/v1/P19-1356} {What does {BERT} learn about the structure of language?}
\newblock In \emph{Proceedings of the 57th Annual Meeting of the Association for Computational Linguistics}, pages 3651--3657, Florence, Italy. Association for Computational Linguistics.

\bibitem[{Kojima et~al.(2022)Kojima, Gu, Reid, Matsuo, and Iwasawa}]{kojima2022large}
Takeshi Kojima, Shixiang~Shane Gu, Machel Reid, Yutaka Matsuo, and Yusuke Iwasawa. 2022.
\newblock Large language models are zero-shot reasoners.
\newblock \emph{Advances in neural information processing systems}, 35:22199--22213.

\bibitem[{Koncel-Kedziorski et~al.(2015)Koncel-Kedziorski, Hajishirzi, Sabharwal, Etzioni, and Ang}]{koncel-kedziorski-etal-2015-parsing}
Rik Koncel-Kedziorski, Hannaneh Hajishirzi, Ashish Sabharwal, Oren Etzioni, and Siena~Dumas Ang. 2015.
\newblock \href {https://doi.org/10.1162/tacl_a_00160} {Parsing algebraic word problems into equations}.
\newblock \emph{Transactions of the Association for Computational Linguistics}, 3:585--597.

\bibitem[{Kopiczko et~al.(2023)Kopiczko, Blankevoort, and Asano}]{kopiczko2023vera}
Dawid~Jan Kopiczko, Tijmen Blankevoort, and Yuki~Markus Asano. 2023.
\newblock Vera: Vector-based random matrix adaptation.
\newblock \emph{arXiv preprint arXiv:2310.11454}.

\bibitem[{Li and Liang(2021)}]{li-liang-2021-Prefix}
Xiang~Lisa Li and Percy Liang. 2021.
\newblock \href {https://doi.org/10.18653/v1/2021.acl-long.353} {Prefix-tuning: Optimizing continuous prompts for generation}.
\newblock In \emph{Proceedings of the 59th Annual Meeting of the Association for Computational Linguistics and the 11th International Joint Conference on Natural Language Processing (Volume 1: Long Papers)}, pages 4582--4597, Online. Association for Computational Linguistics.

\bibitem[{Ling et~al.(2017)Ling, Yogatama, Dyer, and Blunsom}]{ling-etal-2017-program}
Wang Ling, Dani Yogatama, Chris Dyer, and Phil Blunsom. 2017.
\newblock \href {https://doi.org/10.18653/v1/P17-1015} {Program induction by rationale generation: Learning to solve and explain algebraic word problems}.
\newblock In \emph{Proceedings of the 55th Annual Meeting of the Association for Computational Linguistics (Volume 1: Long Papers)}, pages 158--167, Vancouver, Canada. Association for Computational Linguistics.

\bibitem[{Liu et~al.(2024)Liu, Wang, Yin, Molchanov, Wang, Cheng, and Chen}]{liu2024dora}
Shih-Yang Liu, Chien-Yi Wang, Hongxu Yin, Pavlo Molchanov, Yu-Chiang~Frank Wang, Kwang-Ting Cheng, and Min-Hung Chen. 2024.
\newblock Dora: Weight-decomposed low-rank adaptation.
\newblock \emph{arXiv preprint arXiv:2402.09353}.

\bibitem[{Mihaylov et~al.(2018)Mihaylov, Clark, Khot, and Sabharwal}]{mihaylov2018can}
Todor Mihaylov, Peter Clark, Tushar Khot, and Ashish Sabharwal. 2018.
\newblock Can a suit of armor conduct electricity? a new dataset for open book question answering.
\newblock \emph{arXiv preprint arXiv:1809.02789}.

\bibitem[{Novikova et~al.(2017)Novikova, Du{\v{s}}ek, and Rieser}]{novikova2017e2e}
Jekaterina Novikova, Ond{\v{r}}ej Du{\v{s}}ek, and Verena Rieser. 2017.
\newblock The e2e dataset: New challenges for end-to-end generation.
\newblock \emph{arXiv preprint arXiv:1706.09254}.

\bibitem[{Patel et~al.(2021)Patel, Bhattamishra, and Goyal}]{patel-etal-2021-nlp}
Arkil Patel, Satwik Bhattamishra, and Navin Goyal. 2021.
\newblock \href {https://doi.org/10.18653/v1/2021.naacl-main.168} {Are {NLP} models really able to solve simple math word problems?}
\newblock In \emph{Proceedings of the 2021 Conference of the North American Chapter of the Association for Computational Linguistics: Human Language Technologies}, pages 2080--2094, Online. Association for Computational Linguistics.

\bibitem[{Qin et~al.(2023)Qin, Zhang, Zhang, Chen, Yasunaga, and Yang}]{qin2023chatgpt}
Chengwei Qin, Aston Zhang, Zhuosheng Zhang, Jiaao Chen, Michihiro Yasunaga, and Diyi Yang. 2023.
\newblock Is chatgpt a general-purpose natural language processing task solver?
\newblock \emph{arXiv preprint arXiv:2302.06476}.

\bibitem[{Radford et~al.(2019)Radford, Wu, Child, Luan, Amodei, Sutskever et~al.}]{radford2019language}
Alec Radford, Jeffrey Wu, Rewon Child, David Luan, Dario Amodei, Ilya Sutskever, et~al. 2019.
\newblock Language models are unsupervised multitask learners.
\newblock \emph{OpenAI blog}, 1(8):9.

\bibitem[{Roy and Roth(2016)}]{roy2016solving}
Subhro Roy and Dan Roth. 2016.
\newblock Solving general arithmetic word problems.
\newblock \emph{arXiv preprint arXiv:1608.01413}.

\bibitem[{Sakaguchi et~al.(2021)Sakaguchi, Bras, Bhagavatula, and Choi}]{sakaguchi2021winogrande}
Keisuke Sakaguchi, Ronan~Le Bras, Chandra Bhagavatula, and Yejin Choi. 2021.
\newblock Winogrande: An adversarial winograd schema challenge at scale.
\newblock \emph{Communications of the ACM}, 64(9):99--106.

\bibitem[{Sap et~al.(2019)Sap, Rashkin, Chen, LeBras, and Choi}]{sap2019socialiqa}
Maarten Sap, Hannah Rashkin, Derek Chen, Ronan LeBras, and Yejin Choi. 2019.
\newblock Socialiqa: Commonsense reasoning about social interactions.
\newblock \emph{arXiv preprint arXiv:1904.09728}.

\bibitem[{Tenney et~al.(2019)Tenney, Xia, Chen, Wang, Poliak, McCoy, Kim, Durme, Bowman, Das, and Pavlick}]{47786}
Ian Tenney, Patrick Xia, Berlin Chen, Alex Wang, Adam Poliak, R.~Thomas McCoy, Najoung Kim, Benjamin~Van Durme, Samuel~R. Bowman, Dipanjan Das, and Ellie Pavlick. 2019.
\newblock \href {https://openreview.net/forum?id=SJzSgnRcKX} {What do you learn from context? probing for sentence structure in contextualized word representations}.
\newblock In \emph{International Conference on Learning Representations}.

\bibitem[{Touvron et~al.(2023)Touvron, Lavril, Izacard, Martinet, Lachaux, Lacroix, Rozi{\`e}re, Goyal, Hambro, Azhar et~al.}]{touvron2023llama}
Hugo Touvron, Thibaut Lavril, Gautier Izacard, Xavier Martinet, Marie-Anne Lachaux, Timoth{\'e}e Lacroix, Baptiste Rozi{\`e}re, Naman Goyal, Eric Hambro, Faisal Azhar, et~al. 2023.
\newblock Llama: Open and efficient foundation language models.
\newblock \emph{arXiv preprint arXiv:2302.13971}.

\bibitem[{Valipour et~al.(2022)Valipour, Rezagholizadeh, Kobyzev, and Ghodsi}]{valipour2022dylora}
Mojtaba Valipour, Mehdi Rezagholizadeh, Ivan Kobyzev, and Ali Ghodsi. 2022.
\newblock Dylora: Parameter efficient tuning of pre-trained models using dynamic search-free low-rank adaptation.
\newblock \emph{arXiv preprint arXiv:2210.07558}.

\bibitem[{Valipour et~al.(2023)Valipour, Rezagholizadeh, Kobyzev, and Ghodsi}]{valipour-etal-2023-dylora}
Mojtaba Valipour, Mehdi Rezagholizadeh, Ivan Kobyzev, and Ali Ghodsi. 2023.
\newblock \href {https://doi.org/10.18653/v1/2023.eacl-main.239} {{D}y{L}o{RA}: Parameter-efficient tuning of pre-trained models using dynamic search-free low-rank adaptation}.
\newblock In \emph{Proceedings of the 17th Conference of the European Chapter of the Association for Computational Linguistics}, pages 3274--3287, Dubrovnik, Croatia. Association for Computational Linguistics.

\bibitem[{Wang and Komatsuzaki(2021)}]{wang2021gpt}
Ben Wang and Aran Komatsuzaki. 2021.
\newblock Gpt-j-6b: A 6 billion parameter autoregressive language model.

\bibitem[{Wei et~al.(2022)Wei, Wang, Schuurmans, Bosma, Xia, Chi, Le, Zhou et~al.}]{wei2022chain}
Jason Wei, Xuezhi Wang, Dale Schuurmans, Maarten Bosma, Fei Xia, Ed~Chi, Quoc~V Le, Denny Zhou, et~al. 2022.
\newblock Chain-of-thought prompting elicits reasoning in large language models.
\newblock \emph{Advances in neural information processing systems}, 35:24824--24837.

\bibitem[{Zellers et~al.(2019)Zellers, Holtzman, Bisk, Farhadi, and Choi}]{zellers2019hellaswag}
Rowan Zellers, Ari Holtzman, Yonatan Bisk, Ali Farhadi, and Yejin Choi. 2019.
\newblock Hellaswag: Can a machine really finish your sentence?
\newblock \emph{arXiv preprint arXiv:1905.07830}.

\bibitem[{Zhang et~al.(2023)Zhang, Chen, Bukharin, He, Cheng, Chen, and Zhao}]{zhang2023adaptive}
Qingru Zhang, Minshuo Chen, Alexander Bukharin, Pengcheng He, Yu~Cheng, Weizhu Chen, and Tuo Zhao. 2023.
\newblock Adaptive budget allocation for parameter-efficient fine-tuning.
\newblock \emph{arXiv preprint arXiv:2303.10512}.

\end{thebibliography}

\appendix

\section{Experimental Details}\label{A}

\textbf{Data Usage}: The datasets used in this paper come from the open-source work of previous research papers\cite{hu2023llm,hu2021lora}. For the mathematical reasoning tasks, all six datasets are combined by randomly selecting $80\%$ of each, resulting in a total of 3260 data points for training. Testing is then performed on the remaining data for each dataset. For commonsense inference tasks, 170k version of this work\cite{hu2023llm} are used for training, amalgamating the training datasets from all 8 sub-tasks to create this final training dataset, and testing is conducted on their individual testing dataset for each task. For the tasks above, during training and testing, a prompt is added to the data: 'Below is an instruction that describes a task, paired with an input that provides further context. Write a response that appropriately completes the request.' For the E2E dataset, we directly adopte the training and testing datasets used in this work\cite{hu2021lora}.

\textbf{Hyperparameter Settings}: In addition to the hyperparameters mentioned in the text experiments, all other experimental hyperparameters are consistent with those of the main experiment. The experimental hyperparameters of the main experiments for mathematical reasoning, commonsense inference, and E2E tasks are shown in Tables \ref{tab:hyperparameters-math}, \ref{tab:hyperparameters-commonsense}and \ref{tab:hyperparameters}, respectively. The hyperparameters for most baseline experiments are based on references from \cite{hu2023llm} and 2\cite{hu2021lora}, along with their provided open-source code.

All of our methods are consistent with the original LoRA method\cite{hu2021lora}, with the added matrices being parallel to the Q and V matrices. 
The random initialization mentioned in our method follows the Kaiming uniform approach.

\begin{table*}[h]
\centering
\begin{tabular}{lcccccc}
\hline
\textbf{Hyperparameters} & \textbf{Prefix} & \textbf{LoRA} & \textbf{Adapter} & \textbf{Parallel} & \textbf{SARA} & \textbf{Mo-SARA}\\
\hline
% & \textbf{r=16} & \textbf{r=32} & \textbf{r=16} & \textbf{r=32} \\
% \hline
\multicolumn{7}{c}{\textbf{LLaMA-7B}} \\
\hline
Rank $r$ & - & 10 & - & - & - & - \\
$\lambda$ & - & 2 & - & - & - & - \\
Virtual Tokens & 30 & - & - & - & - & - \\
Bottleneck Size & - & - & 256 & 256 & - & -  \\
Threshold & - & - & - & - & 0.01 & 0.5 \\
Parallel Heads & -  & - & - & - & -& 5 \\
Dropout & \multicolumn{6}{c}{0.05} \\
Optimizer & \multicolumn{6}{c}{AdamW} \\
LR & 3e-2 & 3e-4 & 3e-4 & 3e-4 & 3e-3 & 3e-2 \\
LR Scheduler & \multicolumn{6}{c}{Linear} \\
Batch size & \multicolumn{6}{c}{16} \\
Warmup Steps & \multicolumn{6}{c}{100} \\
Epochs & \multicolumn{6}{c}{3} \\
Training Seed & \multicolumn{6}{c}{42} \\
\hline
\multicolumn{7}{c}{\textbf{LLaMA-13B}} \\
\hline
Rank $r$ & - & 10 & - & - & - & - \\
$\lambda$ & - & 2 & - & - & - & - \\
Virtual Tokens & 30 & - & - & - & - & - \\
Bottleneck Size & - & - & 256 & 256 & - & -  \\
Threshold & - & - & - & - & 0.009 & 0.5 \\
Parallel Heads & -  & - & - & - & -& 5 \\
Dropout & \multicolumn{6}{c}{0.05} \\
Optimizer & \multicolumn{6}{c}{AdamW} \\
LR & 3e-2 & 3e-4 & 3e-4 & 3e-4 & 1e-2 & 3e-2 \\
LR Scheduler & \multicolumn{6}{c}{Linear} \\
Batch size & \multicolumn{6}{c}{16} \\
Warmup Steps & \multicolumn{6}{c}{100} \\
Epochs & \multicolumn{6}{c}{3} \\
Training Seed & \multicolumn{6}{c}{42} \\
\hline
\multicolumn{7}{c}{\textbf{GPT-J-6B}} \\
\hline
Rank $r$ & - & 10 & - & - & - & - \\
$\lambda$ & - & 2 & - & - & - & - \\
Virtual Tokens & 30 & - & - & - & - & - \\
Bottleneck Size & - & - & 256 & 256 & - & -  \\
Threshold & - & - & - & - & 0.009 & 0.5 \\
Parallel Heads & -  & - & - & - & -& 5 \\
Dropout & \multicolumn{6}{c}{0.05} \\
Optimizer & \multicolumn{6}{c}{AdamW} \\
LR & 3e-2 & 3e-4 & 3e-4 & 3e-4 & 3e-3 & 3e-2 \\
LR Scheduler & \multicolumn{6}{c}{Linear} \\
Batch size & \multicolumn{6}{c}{16} \\
Warmup Steps & \multicolumn{6}{c}{100} \\
Epochs & \multicolumn{6}{c}{3} \\
Training Seed & \multicolumn{6}{c}{42} \\
\hline
\end{tabular}
\caption{Hyperparameters for Mathematical Reasoning Tasks}
\label{tab:hyperparameters-math}
\end{table*}

\begin{table*}[h!]
\centering
\begin{tabular}{lcccc}
\hline
\textbf{Hyperparameters} & \multicolumn{2}{c}{\textbf{LLaMA-7B}} & \multicolumn{2}{c}{\textbf{LLaMA-13B}} \\
\hline
& \textbf{SARA} & \textbf{Mo-SARA} & \textbf{SARA} & \textbf{Mo-SARA} \\
\hline
Threshold & 0.09 & 0.8 & 0.075 & 0.5 \\
Parallel Heads & - & 5 & - & 5 \\
% $\alpha$ & 32 & 64 & 32 & 64 \\
Dropout & \multicolumn{4}{c}{0.05} \\
Optimizer & \multicolumn{4}{c}{AdamW} \\
LR & 1e-3 & 3e-2 & 1e-3 & 3e-2 \\
LR Scheduler & \multicolumn{4}{c}{Linear} \\
Batch size & \multicolumn{4}{c}{16} \\
Warmup Steps & \multicolumn{4}{c}{100} \\
Epochs & \multicolumn{4}{c}{3} \\
% \hline
% \multicolumn{5}{l}{Where Q,K,V,Up,Dow} \\
\hline
\end{tabular}
\caption{Hyperparameters for Commensense Inference Tasks}
\label{tab:hyperparameters-commonsense}
\end{table*}

\begin{table*}[h!]
\centering
\begin{tabular}{lccc}
\hline
\textbf{Hyperparameters} & \textbf{VeRA} & \textbf{SARA}& \textbf{Mo-SARA} \\
\hline
% & \textbf{VeRA} & \textbf{SARA}& \textbf{Mo-SARA} & \textbf{VeRA} & \textbf{SARA}& \textbf{Mo-SARA} \\
% \hline
Threshold & - & 0.012 & 0.5 \\
Parallel Heads & - & - & 3 \\
% $\alpha$ & 32 & 64 & 32 & 64 \\
Optimizer & \multicolumn{3}{c}{AdamW} \\
LR & 1e-1 & 8e-3 & 7e-2 \\
LR Scheduler & \multicolumn{3}{c}{Linear} \\
Batch size & \multicolumn{3}{c}{16} \\
Weight Decay &\multicolumn{3}{c}{0.01}\\
Lable Smooth &\multicolumn{3}{c}{0.1}\\
Rank & 1024 & - & - \\
LoRA $\alpha$ & 1024 & - & - \\
Warmup Steps & \multicolumn{3}{c}{500} \\
Epochs & \multicolumn{3}{c}{5} \\
Training Seed & \multicolumn{3}{c}{314} \\
% \hline
% \multicolumn{5}{l}{Where Q,K,V,Up,Dow} \\
\hline
\end{tabular}
\caption{Hyperparameters for E2E Task}
\label{tab:hyperparameters}
\end{table*}

\textbf{Model Usage}: In this paper, we utilize four models: LLaMA-7B/13B \cite{touvron2023llama}, GPTJ-6B \cite{wang2021gpt}, and GPT-2\cite{radford2019language}. All training and testing experiments are conducted using a single Nvidia A40, Nvidia RTX4090 or NVIDIA L20.

\section{Relationship between Layers and k under Different Thresholds.}\label{B}

\begin{figure}[!h]
\begin{center}
\includegraphics[scale=0.05]{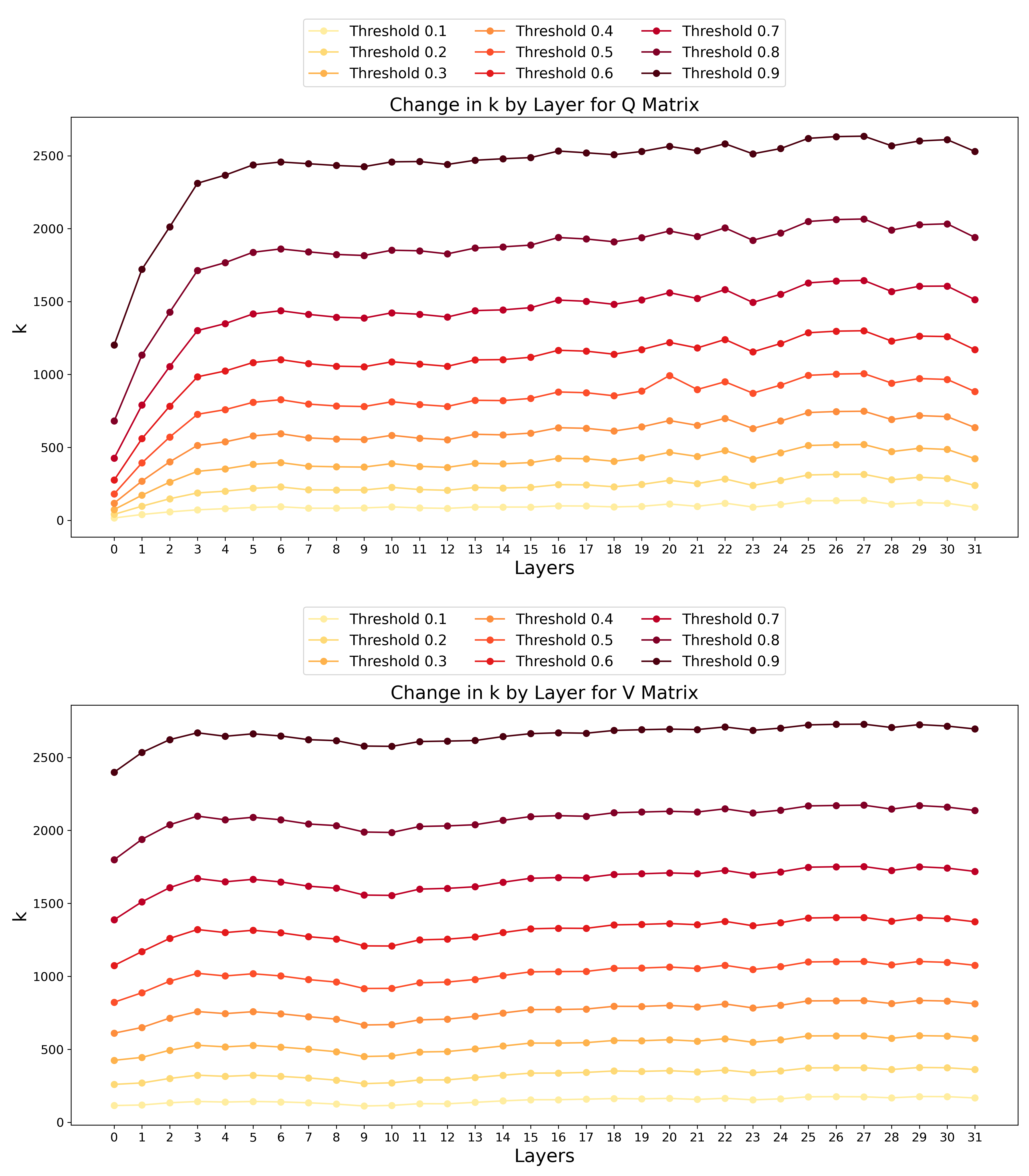}
\caption{Average accuracy of the LoRA method on mathematical reasoning tasks at different $\lambda$ scaling ratios compared to the SARA method.}
\label{fig.8}
\end{center}
\end{figure}

We follow the method described in section \ref{section3} to calculate the k-values obtained from matrix SVD decomposition under different thresholds ranging from 0.1 to 0.9, observing the impact as the number of layers changes. The results for the Q and V matrices are shown in Figures \ref{fig.8}, respectively. All k-values show a trend of initially decreasing around the eighth layer and then increasing as the model's depth increases, which is the opposite of the model performance trend with layer variation, consistent with what we mentioned in section \ref{section3}.

\section{Analysis of the LoRA Method under Different $\lambda$ Hyperparameters.}\label{C}

\begin{figure}[!h]
\begin{center}
\includegraphics[scale=0.45]{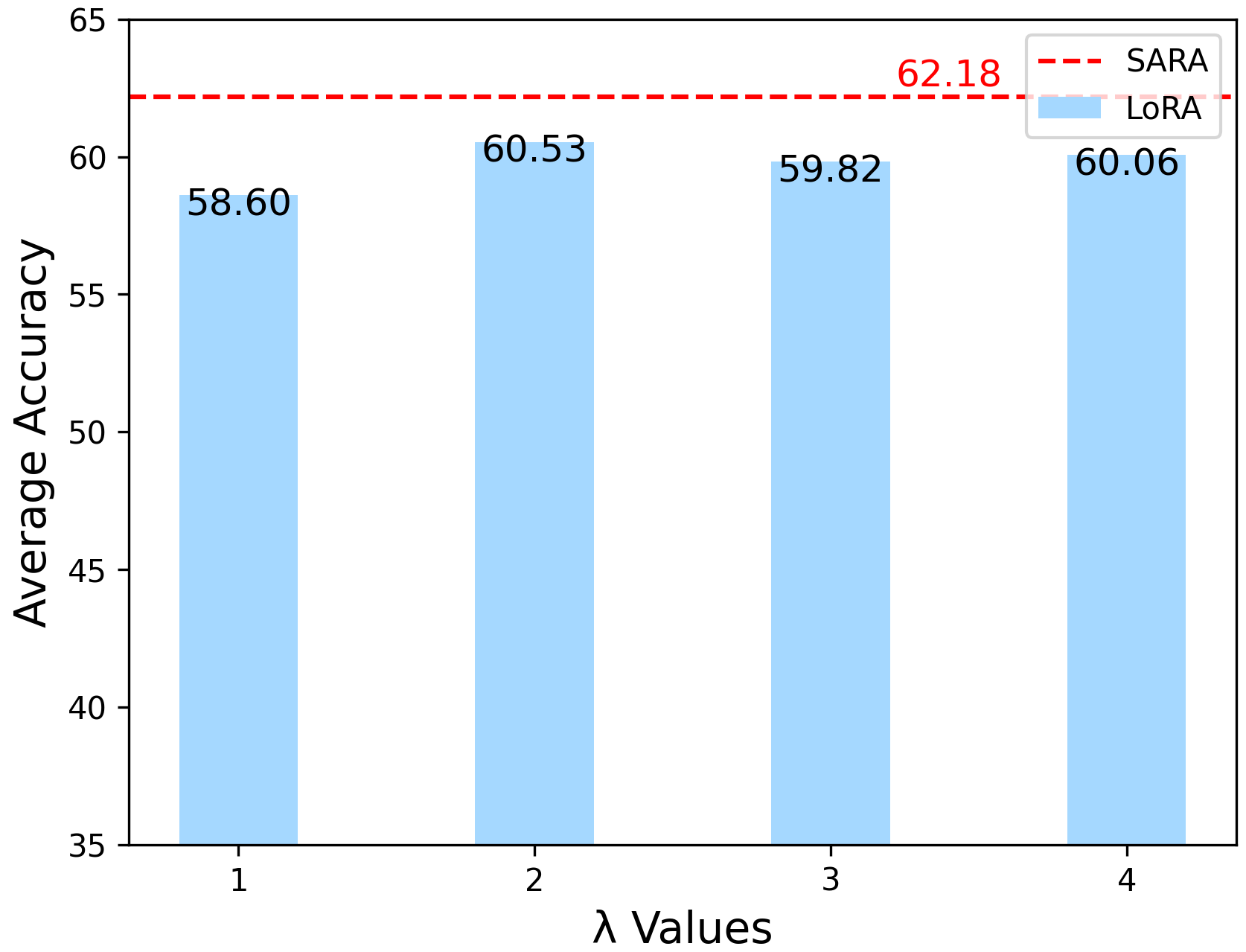}
\caption{Average accuracy of the LoRA method on mathematical reasoning tasks at different $\lambda$ values compared to the SARA method.}
\label{fig.9}
\end{center}
\end{figure}

We modify the $\lambda$ values in the LoRA method into four sets and conduct experiments using LLaMA-7B on the mathematical reasoning tasks. The experimental results are shown in the figure\ref{fig.9}. The original LoRA method is also sensitive to the $\lambda$ hyperparameter values, yielding different results under the four different settings, all of which are lower than those obtained by our SARA method. This indicates that the LoRA method requires validation to find the optimal $\lambda$ values for different tasks, while our approach, which replaces scaling with singular values, partially addresses this issue for adding singular values allows for a more fine-grained determination of the appropriate scaling factor

\section{Heatmaps of routing across layers for various test tasks using the Mo-SARA method.}\label{D}
The experiments for obtaining this heatmaps is consistent with that described in Section \ref{section5.8} of the paper.

% The results from the following 14 figures from Figure \ref{fig.10} to Figure \ref{fig.23} show that mathematical reasoning tasks and commonsense inference tasks exhibit similar routing distributions respectively, and for each layer, there is typically a predominant routing value. This indicates that different sets of singular values play similar roles across different test sets for models trained on the same training set, with each layer being dominated by a specific set of singular values.

The results from the following figures from Figure \ref{fig.10} to Figure \ref{fig.22} show that mathematical reasoning tasks and commonsense inference tasks exhibit similar routing distributions respectively, and for each layer, there is typically a predominant routing value. This indicates that different sets of singular values play similar roles across different test sets for models trained on the same training set, with each layer being dominated by a specific set of singular values.

\section{Supplementary Results for Each Dataset.}\label{E}

Specific results of the experimental supplements on each dataset are presented in the following table\ref{table.8},\ref{table.9},\ref{table.10},\ref{table.11},\ref{table.12} as shown.

\section{Scientific Artifacts}
The datasets we use include the mathematical reasoning dataset SVAMP \citep{patel-etal-2021-nlp},  AQuA \citep{ling-etal-2017-program}, AddSub \citep{hosseini-etal-2014-learning}, MultiArith \citep{roy2016solving}, the SingleEQ \citep{koncel-kedziorski-etal-2015-parsing}, GSM8K \citep{cobbe2021training}, and the commonsense inference dataset ARC \citep{clark2018think}, Boolq \citep{clark-etal-2019-boolq}, WinoGrande \citep{sakaguchi2021winogrande}, PIQA \citep{bisk2020piqa}, SIQA \citep{sap2019socialiqa}, and OBQA \citep{mihaylov2018can}. The pre-trained models we utilize are LLaMA-7B/13B \cite{touvron2023llama}, and GPT-J-6B \cite{wang2021gpt}, as well as E2E task\cite{novikova2017e2e}. All the aforementioned datasets and models are open-source, and our work is solely for scientific research purposes, aligning with their original intent.

\begin{table*}
\centering
\fontsize{9.5pt}{10.5pt}\selectfont
\begin{tabular}{p{2cm}p{1.2cm}p{1.2cm}p{1.2cm}p{1.2cm}p{1.2cm}p{1.2cm}p{1.2cm}}
\hline
\textbf{Method}&\textbf{SVAMP}&\textbf{AQuA}&\textbf{AddSub}&\textbf{MultiArith}&\textbf{SingleEQ}&\textbf{GSM8K}&\textbf{Avg.} \\
\hline
\textbf{LoRA(0-7)} & 48.50& 11.76& 73.42& 74.55& 79.41& 19.32& 51.16
\\
\textbf{LoRA(8-15)} & 49.50&	25.49&	69.62&	84.55&	85.29&	20.45&	55.82
\\
\textbf{LoRA(16-23)} & 40.50& 25.49& 	69.62& 70.00& 76.47& 14.39& 49.41
\\
\textbf{LoRA(24-31)} & 30.50& 25.49& 62.03& 42.73& 61.76& 9.09& 38.60
\\
\hline
\textbf{SARA(0-7)} & 56.00&	29.41&	73.42&	71.82&	82.35&	21.59&	55.77
\\
\textbf{SARA(8-15)} & 54.00&	27.45&	74.68&77.27&	82.35&	22.73&	56.41
\\
\textbf{SARA(16-23)} & 43.50&	31.37&	74.68&	82.73&73.53&	13.64&	53.24
\\
\textbf{SARA(24-31)} & 37.00&15.69&68.35&	64.55&	63.73&	6.06&	42.56
\\
\hline
\end{tabular}
\caption{\label{citation-guide} Supplement to the average accuracy of SARA and LoRA methods across different layers in mathematical reason-
ing tasks(Figure \ref{fig.3}).}
\label{table.8}
\end{table*}

\begin{table*}
\centering
\fontsize{9.5pt}{10.5pt}\selectfont
\begin{tabular}{p{2cm}p{1.2cm}p{1.2cm}p{1.2cm}p{1.2cm}p{1.2cm}p{1.2cm}p{1.2cm}}
\hline
\textbf{Method}&\textbf{SVAMP}&\textbf{AQuA}&\textbf{AddSub}&\textbf{MultiArith}&\textbf{SingleEQ}&\textbf{GSM8K}&\textbf{Avg.} \\
\hline
\textbf{LoRA(r=5)} & 51.50& 23.53& 73.42& 90.91& 87.25& 23.48& 58.35
\\
\textbf{LoRA(r=10)} & 58.50& 23.53& 75.95& 92.73& 88.24& 24.24& 60.53
\\
\textbf{LoRA(r=15)} & 60.00&	17.65&	78.48&	93.64&	86.27&	23.86&	59.98
\\
\textbf{LoRA(r=20)} & 58.50&	19.61&	79.75&	89.09&	87.25&	26.14&	60.06
\\
\hline
\textbf{SARA(0.006)} & 55.00&	19.61&	74.68&	85.45&	88.24&	28.41&	58.57
\\
\textbf{SARA(0.01)} & 60.00&	35.29&	79.75&	89.09&	84.31&	24.62&	62.18
\\
\textbf{SARA(0.016)} & 61.50&	23.53&	78.48&	89.09&	82.35&	26.52&	60.24
\\
\textbf{SARA(0.02)} & 59.50&	25.49&	82.28&	85.45&	84.31&24.62&	60.28
\\
\hline
\end{tabular}
\caption{\label{citation-guide} Supplement to the average accuracy of the SARA and LoRA methods on mathematical reasoning tasks with different trainable parameter counts.(Figure \ref{fig.6})}
\label{table.9}
\end{table*}

\begin{table*}
\centering
\fontsize{9.5pt}{10.5pt}\selectfont
\begin{tabular}{p{2cm}p{1.2cm}p{1.2cm}p{1.2cm}p{1.2cm}p{1.2cm}p{1.2cm}p{1.2cm}}
\hline
\textbf{Method}&\textbf{SVAMP}&\textbf{AQuA}&\textbf{AddSub}&\textbf{MultiArith}&\textbf{SingleEQ}&\textbf{GSM8K}&\textbf{Avg.} \\
\hline
\textbf{Threshold=0.1} & 51.50& 	27.45& 	69.62& 84.55& 82.35& 23.86& 56.56
\\
\textbf{Threshold=0.3} & 55.00&	25.49&	77.22& 85.45&	82.35&23.86&58.23
\\
\textbf{Threshold=0.5} & 56.00&	23.53&	73.42&	89.09&	84.31&	23.48&	58.31
\\
\textbf{Threshold=0.7} & 56.50&	15.69&	73.42&	90.91&	85.29&	23.86&	57.61
\\
\hline
\end{tabular}
\caption{\label{citation-guide} Supplement to the average accuracy of Mo-SARA (1 head) on mathematical reasoning tasks under different thresholds.(Figure\ref{fig.4})}
\label{table.10}
\end{table*}

\begin{table*}
\centering
\fontsize{9.5pt}{10.5pt}\selectfont
\begin{tabular}{p{3cm}p{1.2cm}p{1.2cm}p{1.2cm}p{1.2cm}p{1.2cm}p{1.2cm}p{1.2cm}}
\hline
\textbf{Method}&\textbf{SVAMP}&\textbf{AQuA}&\textbf{AddSub}&\textbf{MultiArith}&\textbf{SingleEQ}&\textbf{GSM8K}&\textbf{Avg.} \\
\hline
\textbf{Mo-SARA(1 head)} & 56.00&	23.53&	73.42&	89.09&	84.31&	23.48&	58.31
\\
\textbf{Mo-SARA(3 head)} & 54.50& 21.57& 75.95& 89.09& 85.29& 23.86& 58.38
\\
\textbf{Mo-SARA(5 head)} & 55.00&	23.53&	70.89&	90.91&	87.25&	26.14&	58.95
\\
\textbf{Mo-SARA(7 head)} & 55.50& 23.53& 	75.95& 90.00& 85.29& 25.00& 59.21 \\
\textbf{Mo-SARA(9 head)} & 53.00 & 25.49 & 78.48 & 88.18 & 86.27 & 24.62 & 59.34
\\
\hline
\end{tabular}
\caption{\label{citation-guide} Supplement to the average accuracy of Mo-SARA on mathematical reasoning tasks with different numbers of parallel heads, compared to SARA and LoRA methods.(Figure \ref{fig.5})}
\label{table.11}
\end{table*}

% \subsection{Analysis of the Gating Mechanism}\label{D.3}
% To further investigate the impact of our joint Lora method, we visualize the linear layer used to generate the gate next to the Q and V matrices in the llama-7b model trained on mathematical task. The experimental results, shown in Figure \ref{fig.3}, reveal varying weights across different layers, explaining how our approach assists the model in allocating different weights at different layers for improved performance and enhanced joint fine-tuning capabilities across tasks. We believe the differences between parameters may mainly result from the mechanism used to handle diverse inputs across different layers and enhance model performance

\begin{table*}
\centering
\fontsize{9.5pt}{10.5pt}\selectfont
\begin{tabular}{p{1.2cm}p{1.2cm}p{1.2cm}p{1.2cm}p{1.2cm}p{1.2cm}p{1.2cm}p{1.2cm}}
\hline
\textbf{Method}&\textbf{SVAMP}&\textbf{AQuA}&\textbf{AddSub}&\textbf{MultiArith}&\textbf{SingleEQ}&\textbf{GSM8K}&\textbf{Avg.} \\
\hline
\textbf{$\lambda$=1} & 52.50&	23.53&	74.68&	90.91&	87.25& 22.73&	58.60
\\
\textbf{$\lambda$=2} & 58.50& 23.53& 75.95& 92.73& 88.24& 24.24& 60.53
\\
\textbf{$\lambda$=3} & 58.00&	19.61&	74.68&	93.64&	87.25&	25.76&	59.82
\\
\textbf{$\lambda$=4} & 58.00&	21.57&	74.68&	93.64&88.24&	24.24&	60.06
\\
\hline
\end{tabular}
\caption{\label{citation-guide} Supplement to the average accuracy of the LoRA method on mathematical reasoning tasks at different $\lambda$ values(Figure \ref{fig.9})}
\label{table.12}
\end{table*}

\end{document}